\documentclass[10pt,twocolumn,letterpaper]{article}

\usepackage{cvpr}      

\usepackage{algorithm}
\usepackage{algorithmicx}
\usepackage{algpseudocode}

\usepackage{amsmath}
\usepackage{amssymb}
\usepackage{wasysym}
\usepackage{graphicx}
\usepackage{pifont}  
\usepackage{multirow}   

\newcommand{\method}{Bearing-UAV\xspace}
\newcommand{\dataset}{\method-90K\xspace}
\newcommand{\grid}{M2T\xspace}
\newcommand{\supp}{Suppl.~}

\usepackage{microtype}

\renewcommand{\paragraph}[1]{\vspace{.5em}\noindent\textbf{#1.}}

\usepackage{bm}

\definecolor{cvprblue}{rgb}{0.21,0.49,0.74}
\usepackage[pagebackref,breaklinks,colorlinks,allcolors=cvprblue]{hyperref}

\usepackage[capitalize]{cleveref}
\crefname{section}{Sec.}{Secs.}
\Crefname{section}{Section}{Sections}
\Crefname{table}{Table}{Tables}
\crefname{table}{Tab.}{Tabs.}

\def\paperID{9732} 
\def\confName{CVPR}
\def\confYear{2026}

\title{Beyond Matching to Tiles: Bridging Unaligned Aerial and Satellite Views for Vision-Only UAV Navigation}

\author{
\textbf{Kejia Liu$^{1 *}$, Haoyang Zhou$^{1 *}$, Ruoyu Xu$^{1 *}$, Peicheng Wang$^{1}$\thanks{These authors contributed equally.} , Mingli Song$^{1,2,3}$, Haofei Zhang$^{2,3}$ \thanks{Corresponding author, email: haofeizhang@zju.edu.cn}}\\
$^1$College of Computer Science and Technology, Zhejiang University,\\
$^2$State Key Laboratory of Blockchain and Data Security, Zhejiang University,\\
$^3$Hangzhou High-Tech Zone (Binjiang) Institute of Blockchain and Data Security\\
}

\begin{document}
\maketitle

\begin{abstract}

Recent advances in cross-view geo-localization (CVGL) methods have shown strong potential for supporting unmanned aerial vehicle (UAV) navigation in GNSS-denied environments.
However, existing work predominantly focuses on matching UAV views to onboard satellite tiles, which introduces an inherent trade-off between accuracy and storage overhead, and overlooks the importance of the UAV's heading during navigation.
Moreover, the substantial discrepancies and varying overlaps in cross-view scenarios have been insufficiently considered, limiting their generalization to real-world scenarios.
In this paper, we present \method, a purely vision-driven cross-view navigation method that jointly predicts UAV absolute location and heading from neighboring features, enabling accurate, lightweight, and robust navigation in the wild.
Our method leverages global and local structural features and explicitly encodes relative spatial relationships, making it robust to cross-view variations, misalignment, and feature-sparse conditions.
We also present \dataset, a multi-city benchmark for evaluating cross-view localization and navigation.
Extensive experiments show encouraging results that \method yields lower localization errors than previous matching/retrieval paradigms across diverse terrains.
Our code is publicly available at \url{https://github.com/liukejia121/bearinguav}.
\end{abstract}

\section{Introduction}
\label{sec:intro}

Recent years have witnessed the widespread deployment of unmanned aerial vehicles (UAVs) across critical domains such as the low-altitude economy~\cite{shihavuddin2019wind, jarraya2025gnssdenied}, emergency response~\cite{rossi2014gasdrone}, and industrial applications~\cite{nooralishahi2021dronebased}.
However, current UAV localization and navigation systems, which rely heavily on wireless signals and manual operation, remain vulnerable to interference and face persistent challenges in ensuring both safety and autonomy~\cite{gurgu2022visionbased,li2023jointly,zhu2023sues200,liu2019cvact,zhu2021vigor}.

\begin{figure}[!t]
\centering%
    \includegraphics[width=\linewidth]{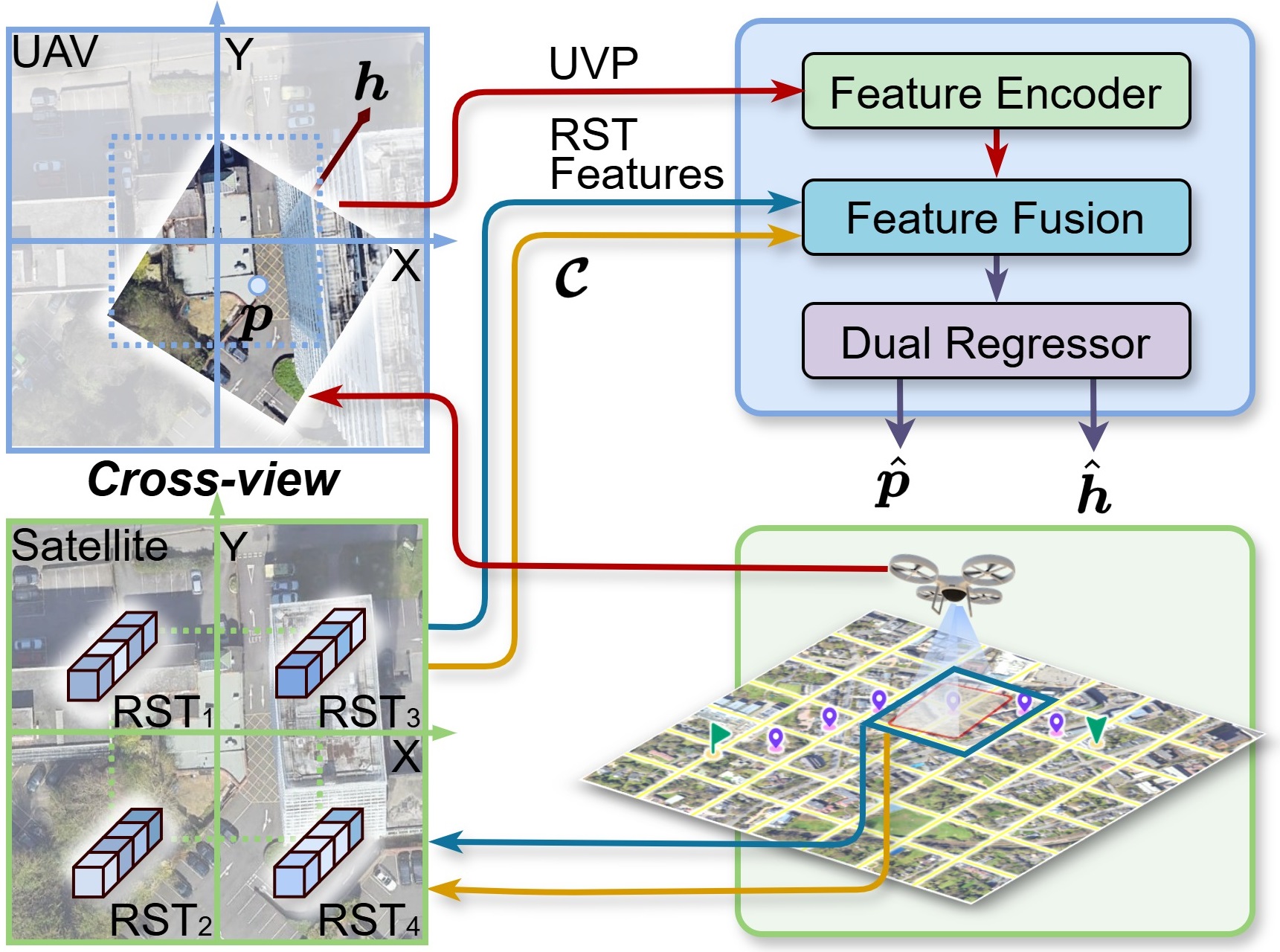}
\caption{\method overview.
Given a UAV-view patch (UVP) and four adjacent remote-sensing tile (RST) features with relative coordinates ($\boldsymbol{\mathcal{C}}$), the model jointly regresses position and heading.}%
\label{fig:intro}%
\end{figure}

Cross-view geo-localization (CVGL), a set of purely vision-based UAV navigation approaches~\cite{klammer2024bevloc, durgam2024crossview, couturier2024review, masselli2016localization}, has been proposed to address these challenges by matching UAV-captured views with geo-referenced satellite tiles encoded by deep models~\cite{xiao2025visionbased, sarlin2023orienternet, zheng2020university1652}.

Current methods following the matching-to-tile (M2T) paradigm fall into two major classes. 
One class of methods predicts the UAV's position by matching UAV views to onboard satellite tiles~\cite{he2025localization, ren2024uavsbased, yao2024uav, mughal2021assisting, masselli2016localization}.
These approaches repeatedly encode tiles, leading to significant computational overhead and carrying complete satellite imagery makes storage scale quadratically.
The other class of approaches pre-encodes satellite tiles into lightweight and discretized feature vectors using deep models~\cite{ji2025game4loc, dai2024denseuav2, zhu2023sues200, zheng2020university1652, zhu2021vigor}.
While conducting similarity search to retrieve current location greatly improves storage and computational efficiency, the localization accuracy is constrained by the grid density.

However, supporting UAV navigation requires not only accurate localization but also reliable heading information, which is largely overlooked by current methods, limiting their ability to drive end-to-end navigation.
Recently, AngleRobust~\cite{wang2024angle} directly predicts azimuth from a sequence of UAV views, but its applicability is confined to a single, densely sampled corridor.
More importantly, existing navigation models are typically trained on datasets that overlook the inherent differences and misalignment between UAV views and satellite tiles, making them difficult to generalize well to real-world scenarios.
Therefore, bridging unaligned aerial-satellite views for vision-only UAV navigation beyond tile matching remains an open problem.

Towards this end, we propose a novel cross-view position-and-heading regression network for learning visual bearing, termed \textbf{\method}, along with a dataset \dataset, containing 90k cross-view image pairs for training and evaluation.
The proposed \method jointly estimates the precise coordinates beyond \grid resolution and the heading angle under cross-view conditions, while remaining robust to variations in misalignment, weather and \grid density, thereby supporting UAV autonomous navigation in the wild.

As shown in~\cref{fig:intro}, \method takes features of four adjacent remote-sensing tiles (RSTs, \ie, satellite-view tiles) and one UAV-view patch (UVP) as inputs, and directly regresses absolute position and heading angle.
Instead of \grid paradigm, which ties localization accuracy to tile density, \method exploits surrounding information to regress UAV position beyond the \grid's resolution.
Furthermore, to bridge the aerial-satellite view gap, we leverage relative positional cues from adjacent tiles to provide localization guidance and employ cross-attention to focus on overlapping regions, hence improving the accuracy of both position and heading regression under misalignment and feature sparsity.
Extensive experiments on our proposed \dataset demonstrate that:
(1) the localization accuracy of \method surpasses existing matching/retrieval paradigms by a large margin;
(2) thanks to the heading branch, \method enables end-to-end navigation with a high success rate under the cross aerial-satellite view condition;
(3) \method remains robust to various weather effects.

Our main contributions are summarized as:
\begin{itemize}
    \item We propose a novel geo-localization paradigm beyond \grid, achieving higher localization accuracy.
    
    \item We introduce a lightweight, multi-task model which enables efficient localization and heading prediction, thereby supporting reliable long-range navigation.

    \item To address viewpoint-induced parallax, misalignment, and feature sparsity in UAV-satellite cross-view settings, we construct \dataset dataset to ensure that our paradigm can be applied to more realistic scenarios.
\end{itemize}

\section{Related Work}
\label{sec:related_work}
As a core component of remote-sensing-based vision navigation, CVGL aims to solve geo-localization between UAV's low-altitude oblique imagery and high-altitude, orthorectified satellite references. A key challenge is the significant viewpoint-induced parallax at the same geographic location due to different viewpoints.

\subsection{Cross-View Geo-Localization}

By enabling cross-view matching between ground-view and satellite-view (G-S) images~\cite{scottworkman2015cvusa,hu2018cvmnet,liu2019cvact,zhu2021vigor,zheng2020university1652}, CVGL has been a very important alternative for localization in GNSS-denied environments~\cite{durgam2024crossview}.
Inspired by advances in G-S cross-view localization, researchers are introducing CVGL for UAV vision-based localization~\cite{couturier2024review}.
University-1652~\cite{zheng2020university1652} constructed and publicly released the first cross-view dataset for UAV and satellite (U-S) and successfully localized buildings from the UAV perspective via feature retrieval, which in turn motivated more UAV CVGL datasets~\cite{zhu2023sues200,dai2024denseuav2,hou2025mcfa} and related algorithms~\cite{yang2025vimgeo,chen2025obtpn,cui2023novel}. 
Further, several works~\cite{zhu2023sues200,dai2024denseuav2,ji2025game4loc} adopt discrete but spatially contiguous satellite tiles to better approximate real scenes. However, this shift exacerbates spatial misalignment and feature sparsity, ultimately degrading retrieval/matching accuracy. 
To address these issues, some approaches improve feature discrimination via feature segmentation~\cite{dai2022transformerbased}, multi-scale feature~\cite{hou2025mcfa}, or local-feature aggregation~\cite{duan2024sgmnet,zhang2025hierarchical}. Others introduce attention mechanisms~\cite{cui2023novel,yang2025vimgeo,chen2025obtpn} or new models and schemes~\cite{yang2025vimgeo,ji2025game4loc}.
At a broader level, CurBench~\cite{zhou2024curbench} and CurML~\cite{zhou2022curml} provide the first benchmark and library for curriculum learning. 
It still remains an open question of robust cross-view localization under misalignment and low feature density, especially with varying IoUs.
To this end, we build an end-to-end network that fuses cross-view features via skip connections and directly regresses UAV position.

\begin{figure*}[!t]
\centering%
    \includegraphics[width=\textwidth]{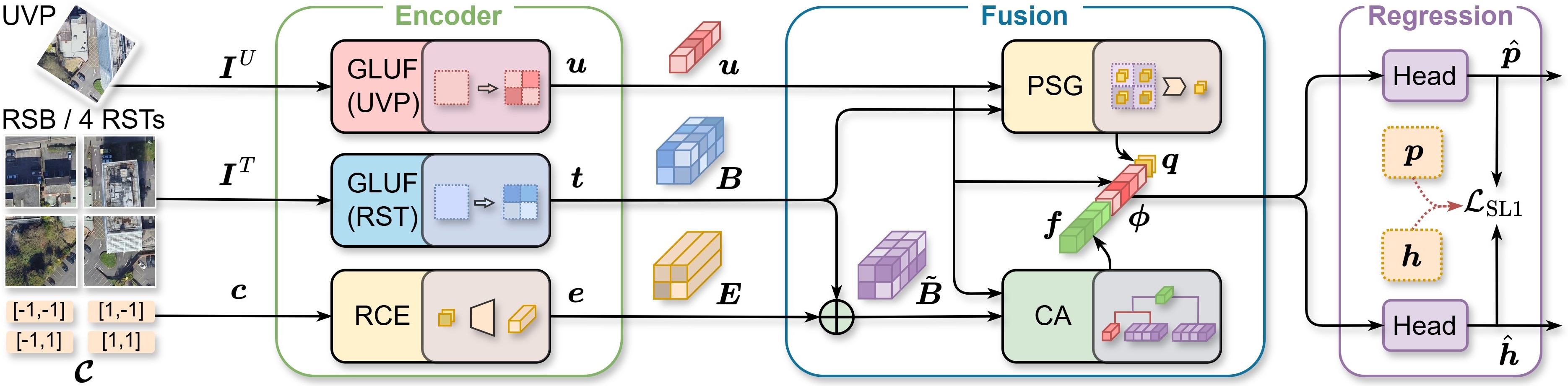}
\caption{\method in training mode. Given four adjacent RSTs with their relative coordinates ($\boldsymbol{\mathcal{C}}$) and one UVP within the same remote-sensing block (RSB), 
\method predicts the UAV’s absolute position and heading angle. 
Two GLUF submodules with shared parameters extract UVP/RST features. 
RCE encodes the relative coordinate of each RST. 
CA captures overlap-aware cross-view correspondences via cross-attention between UVP and RST features.
PSG estimates a weighted position vector by aggregating UVP-RSTs patch similarities.
We then fuse these output features via a residual-style connection, and apply two independent heads to jointly regress position and heading.}
\label{fig:cvphr}%
\end{figure*}

\subsection{Purely Vision-based Orientation Awareness}

\noindent
Most existing CVGL methods focus solely on localization in urban scenes and fail to suppress UAV rotational drift. 
Beyond cross-view appearance gaps and misalignment, heading estimation is further constrained by visual-geometric ambiguities, rotational symmetries, and the lack of an absolute orientation reference. 
Consequently, many works pursue purely vision-based heading perception~\cite{wang2024view,klammer2024bevloc,shi2023boosting,wangfinegrained}. Among them, Wang \textit{et al.} \cite{shi2020where} first achieve G-S cross-view localization and orientation of ground-view images by retrieving the geographic location and then estimating orientation. 
Nevertheless, purely vision-based UAV orientation remains underexplored ~\cite{yang20243d}. Although \cite{shetty2019uava,chen2024realtimeb} report high accuracy, their datasets are idealized and they rely on visual odometry for on-board camera poses. 
Existing purely vision-based approaches are predominantly two-stage: localize first, then orient. 
For instance, methods estimate heading via mutual information~\cite{sun2025pfedcrossview}, motion-matrix rotation, or feature-geometry cues~\cite{sun2025pfedcrossview,qiu2025highprecision}; multi-rotation matching is also common~\cite{wang2025vecmaplocnet}.
These methods require the localization result as a global pose anchor, thereby propagating localization errors to orientation. ~\cite{wang2024absolute} uses single-stage pure vision pose estimation but has large heading errors in cross-view scenarios.
To address this challenge, we augment our model with parallel regression heads that use four adjacent satellite tiles to simultaneously estimate the UAV's position and heading.

Moreover, PnP-based methods estimate 6-DoF poses via geometric solvers~\cite{dhaouadi2025ortholoca, ye2025exploring, schleissvpair}, while sensor-fusion methods use onboard sensors~\cite{chen2024realtimeb, he2023foundloc}. We focus on vision-only U-S cross-view localization rather than G-S setting~\cite{shi2024weakly, wang2023view, shi2022accurate}.
\section{Method}
\label{sec:3_method1_1}

We present Bearing-UAV and its navigation scheme Bearing-Naver in this section.
Acronyms are listed in \supp\ref{sec:acronyms}.

\subsection{Cross-View Position-Heading Regression}  
As shown in~\cref{fig:cvphr}, the overall procedure of Bearing-UAV includes a feature extraction module that extracts comprehensive cross-domain features and positional cues, a fusion module that captures cross-domain correspondences, and dual regression heads that predict position and heading.
We group four adjacent RSTs into a remote-sensing block (RSB).
Leveraging four RSTs per UVP and modeling their interactions via cross-attention and similarity improves robustness to misalignment and sparse-feature conditions.

\subsubsection{Feature Extraction Module}  

\begin{figure}[!t]
\centering%
    \includegraphics[width=\linewidth]{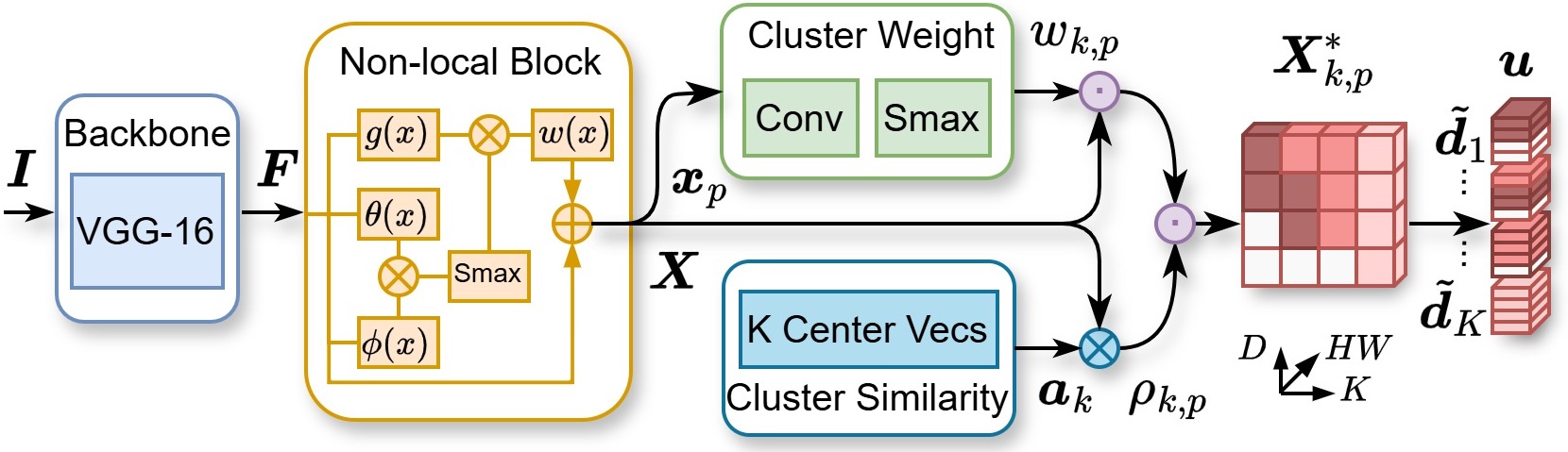}
\caption{Global-Local Unity Feature (GLUF).
A VGG-16 extracts feature map $\bm{F}$ from input image $\bm{I}$, 
a Non-local Block produces semi-global feature map $\bm{X}$ composed of local descriptors $\bm{x}_p$. 
Two clustering branches compute cluster weights and similarities with learnable centers $\bm{a}_k$ to obtain feature $\bm{X}^{*}_{k,p}$, 
which are aggregated across spatial sites and concatenated to form the GLUF vector $\bm{u}$.
}
\label{fig:gluf}%
\end{figure}

Building on the preceding discussion of cross-view localization, we target robustness under cross-view misalignment and varying IoUs.
As shown in~\cref{fig:gluf}, we propose a Global-Local Unity Feature (GLUF) submodule that jointly encodes global contextual similarity and clustered local segments,
enabling correspondences even when cross-domain images overlap only partially. 
Moreover, the Relative Coordinate Encoder (RCE) encodes the relative coordinates of four RSTs into embeddings for the corresponding GLUF vectors.

\paragraph{Global-Local Unity Feature (GLUF)}
To jointly exploit global and local cues for accurate UAV localization, we first extract a feature map using a backbone network (\eg, VGG-16~\cite{Simonyan2015VeryDeepCNN}),
a non-local block~\cite{wang2018nonlocal} is then applied to capture long-range dependencies and enhance local responses. 
Following SGMNet~\cite{duan2024sgmnet}, a clustering scheme generates multiple semi-global descriptors, which are aggregated into a unified representation termed the GLUF vector.
The GLUF-enhanced features provide global similarity cues for inter-tile matching while maintaining ordered, position-aware local feature segments for cross-attention. 
Such patch clustering module (\textit{e.g.}, SGMNet) can be replaced by other suitable feature extractors with only minor performance degradation.

Let $\bm{F}\in\mathbb{R}^{H\times W\times D} = \operatorname{backbone}(\bm{I})$ be the encoded feature of the UVP or RST, the semi-global feature map is obtained by a non-local block:
\begin{equation}
    \bm{X}=\mathrm{NLConv}(\bm{F})\in\mathbb{R}^{H\times W\times D} \text{.}
\end{equation}
Let $\Omega=\{(i, j): i=1, \ldots, H, j=1, \ldots, W\}$ be the spatial sites (\ie, index set) for $\bm{X}$.
For position $p\in\Omega$, let $\bm{x}_p\in\mathbb{R}^{D}$ be the \emph{site-level local descriptor}, and let $\{\bm{a}_k\}_{k=1}^{K}$ be $K$ learnable cluster centers of site-level local features. 
Then, the cluster weight $\bm{w}_{p}$ determines the assignment of $\bm{x}_p$ to the cluster centers ${\bm{a}_k}$ as follows:
\begin{equation}
\bm{w}_{p}=\operatorname{softmax} \left(\left[\bm{a}_k^\top\bm{x}_p-\|\bm{a}_k\|_2 \right]_{k=1}^K\right) \text{.}
\end{equation}
The similarity score $\rho_{k,p}=\operatorname{ReLU}(\bm{a}_k^\top\bm{x}_p)$  measures the affinity between $\bm{x}_p$ and the center vector $\bm{a}_k$.
The clustered feature $\bm{X}^*\in\mathbb{R}^{K\times D\times HW}$ can be calculated by:
\begin{equation}
\bm{X}^{*}_{k,p}=w_{k,p}\,\rho_{k,p}\,\bm{x}_p \text{.}
\end{equation}
Aggregating $\bm{X}^{*}_{k,p}\in\mathbb{R}^{D}$ over space $\Omega$ yields the \emph{cluster-level local features} $\bm{d}_k\in\mathbb{R}^{D}$ and its normalized vector:
\begin{equation}
\bm{d}_k=\sum\nolimits_{p\in\Omega}\bm{X}^{*}_{k,p},\quad
\tilde{\bm{d}}_{k}=\frac{\bm{d}_k}{\lVert\bm{d}_k\rVert_2}.
\end{equation}
Concatenating $K$ normalized cluster descriptors and re-normalizing gives the GLUF vector $\bm{u}$:
\begin{equation}
\bm{u} := \operatorname{norm} \left([\tilde{\bm{d}}_{1}:\tilde{\bm{d}}_{K}]\right) \in\mathbb{R}^{KD}\text{.}
\end{equation}

\paragraph{Relative Coordinate Encoder (RCE)}  
\method regresses the UVP's position and heading from four adjacent RSTs.
RCE is a lightweight multilayer perceptron (MLP) of depth $L$, the dimension of each layer is $[d_1,\dots,d_L]$ with $d_0=2$ and $d_L=KD$.
Let $\bm{\mathcal{C}}=\{\bm{c}_j\}_{j=1}^4$, $\bm{c}_j\in\mathbb{R}^2$, denote the 2D relative
coordinates of the four RSTs \wrt the RSB center, 
where $\bm{y}_j^{(0)}=\bm{c}_j$ and $\sigma(\cdot)=\mathrm{ReLU}(\cdot)$.
The layer $\ell$ of RCE computes:
\begin{equation}
\bm{y}_j^{(\ell)} := \sigma\left(\bm{W}_\ell^{RCE}\bm{y}_j^{(\ell-1)}+\bm{b}_\ell^{RCE}\right) \text{,}
\end{equation}
where $\bm{W}_\ell^{RCE}\in\mathbb{R}^{D_\ell\times D_{\ell-1}}$, $\bm{b}_\ell^{RCE}\in\mathbb{R}^{D_\ell}$.
The coordinate embedding is the output from the final layer $\bm{e}_j=\bm{y}_j^{(L)}$.

\subsubsection{Cross-View Feature Fusion Module}  
\label{sec:3_method1_2}

The feature fusion module first injects tile positional cues for fusion using a ViT-style positional embedding~\cite{Dosovitskiy2021animage}, it then extracts cross-view feature via Cross-Attention (CA) submodule and estimates a similarity-weighted guidance coordinate using the Patch Similarity-Guided (PSG) submodule. Finally, the cross-correlation feature, UVP descriptor, and guidance coordinate are concatenated to construct the fused representation for prediction.

Let the GLUF vector of UVP be $\bm{u}\in\mathbb{R}^{KD}$, 
the GLUF vectors $\{\bm{t}_j\in\mathbb{R}^{KD}\}_{j=1}^4$ of the four RSTs form a tensor $\bm{B}\in\mathbb{R}^{2\times2\times KD}$, 
and the corresponding four relative-coordinate embeddings $\{\bm{e}_j\}_{j=1}^4$ form a tensor $\bm{E}\in\mathbb{R}^{2\times2\times KD}$.
Since the GLUF vector is already normalized, we simply define $\tilde{\bm{B}} := \bm{B}+\bm{E}$ as the position-injected RST features, which expose relative layout to the fusion stage and helps the network learn position-angle relationships under supervision.

\paragraph{Patch Similarity-Guided (PSG)}
Under our regression setting, the UVP mostly overlaps four adjacent RSTs. 
We leverage the U-S cross-view cosine similarity between the UVP and RSTs to compute a weighted guidance coordinate, 
providing a strong prior for position regression by emphasizing RST regions corresponding to the UVP location.

Reshape $\tilde{\bm{B}}$ into $\{\tilde{\bm{b}}_{j}\in\mathbb{R}^{KD}\}_{j=1}^4$, then we compute cosine similarities across the four neighbors:
\begin{equation}
\bm{\alpha} = \operatorname{softmax}\left(\left[\cos(\bm{u},\tilde{\bm{b}}_{j})\right]_{j=1}^4\right) \in \mathbb{R}^{4} \text{.}
\end{equation}
Then, by taking a weighted sum of the RSTs' relative coordinates $\bm{c}_j$, we obtain the similarity-guided coordinate:
\begin{equation}
\bm{q} := \sum\nolimits_{j=1}^{4} \alpha_{j}\bm{c}_j \in \mathbb{R}^{2} \text{.}
\end{equation}

\paragraph{Cross-Attention (CA)}
As the UVP generally overlaps the four RSTs to different degrees, 
we apply lightweight cross-attention submodule to extract overlap-aware associations, enabling the model to learn essential cross-view correspondences between the UVP and RSTs under misalignment and sparse-feature conditions. 
Thus, the resulting cross-view features support the joint regression of position and heading.

Let $\bm{Q}=\bm{W}^{Q} \bm{u}\in\mathbb{R}^{d}$, $\bm{K}=[\bm{W}^K \tilde{\bm{b}}_{1}, \ldots, \bm{W}^K \tilde{\bm{b}}_{4}]\in\mathbb{R}^{4\times d}$, and $\bm{V}=[\bm{W}^V \tilde{\bm{b}}_{1}, \ldots, \bm{W}^V \tilde{\bm{b}}_{4}]\in\mathbb{R}^{4\times d}$ be the queries from UVP and keys/values from four adjacent RSTs, accordingly.
The scaled dot-product attention computes the cross-view feature:
\begin{equation}
\bm{f} := \operatorname{softmax} \left(\frac{\bm{Q}\bm{K}^\top}{\sqrt{d}}\right) \bm{V} \in\mathbb{R}^{KD} \text{.}
\end{equation}

The fused feature $\bm{\phi}$ is the concatenated vector from the UVP descriptor $\bm{u}$, the cross-attended feature $\bm{f}$, and the similarity-guided coordinate $\bm{q}$:
\begin{equation}
\bm{\phi} := \operatorname{concat}(\bm{u}, \bm{f}, \bm{q})\in\mathbb{R}^{KD+KD+2} \text{.}
\end{equation}

\subsubsection{Position-Heading Regression Module}  
\label{sec:3_method1_3}

Both regression heads take the same fused feature $\bm{\phi}$ as input.
Each head is an $M$-layer MLP with ReLU activations, where the final layer maps the intermediate features to position coordinate and heading angle.
For the $m$-th layer, position feature $\bm{p}^{(m)}$ and heading feature $\bm{h}^{(m)}$ are computed as:
\begin{equation}
    \begin{aligned}
        \bm{p}^{(m)} & = \sigma \left(\bm{W}_{m}^{PR}\,\bm{p}^{(m-1)}+\bm{b}_{m}^{PR}\right)\\
        \bm{h}^{(m)} & = \sigma \left(\bm{W}_{m}^{HR}\,\bm{h}^{(m-1)}+\bm{b}_{m}^{HR}\right)
    \end{aligned}\text{,}
\end{equation}
where $\bm{p}^{(0)}=\bm{h}^{(0)}=\bm{\phi}$; $\hat{\bm{p}}=\bm{p}^{(M)}\in\mathbb{R}^2$ represents the relative coordinates and $\hat{\bm{h}}=\bm{h}^{(M)}=(\cos\hat{\theta},\sin\hat{\theta})\in\mathbb{R}^2$ represents the heading directional vector.
We would like to point out that the heading angle is represented as a vector rather than a raw angle $\hat{\theta}$ to resolve the periodicity ambiguity and provide a continuous, well-behaved regression target.

\subsection{Bearing-Naver}
\label{sec:3_method2}

\begin{figure}[!t]
\centering%
    \includegraphics[width=\linewidth]{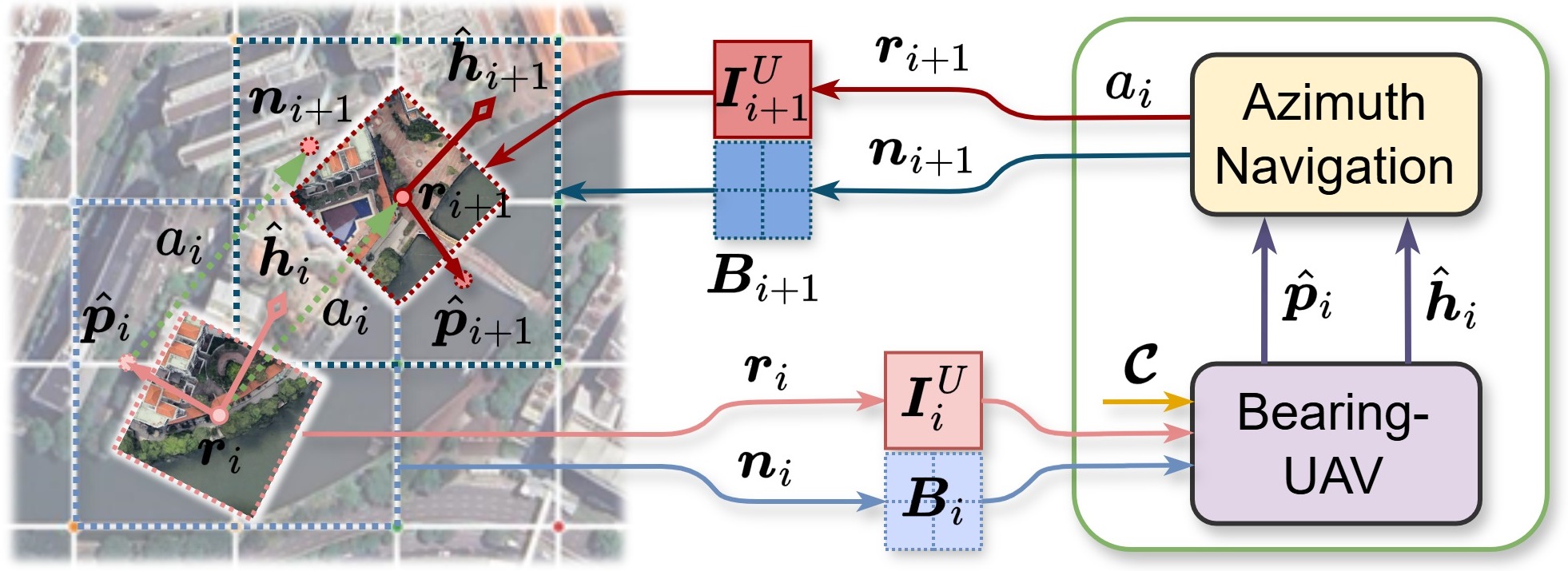}
\caption{Bearing-Naver’s operating mode. At each step, we first collect the UVP and its corresponding RST features, use Bearing-UAV to regress the position and heading, and then compute the azimuth from the current position to the next waypoint, aligning the heading with this azimuth to update UAV state for the next step.}%
\label{fig:naver}%
\end{figure}

By modeling the satellite image as a set of overlapping RSBs and using the proposed Bearing-UAV method, we construct a purely vision-driven point-to-point navigation scheme along specified waypoints in urban scenes, termed \textbf{Bearing-Naver}.
Initialized from a known start position in a certain RSB, this navigation scheme can be summarized as sequentially searching for the next step as shown in~\cref{fig:naver}.
Bearing-Naver supports pre-converting the onboard RSTs into a compact feature table, enabling lightweight and efficient lookup-based UAV flight. During training, the RSTs are instead encoded by the backbone to produce GLUF vectors.

Let $\bm{r}_i\in\mathbb{R}^2$ be the real position at step $i$ and $\bm{n}_i\in\mathbb{R}^2$ be the nominal position (the UAV ``believes'' where it is and indexes RSB with $\bm{n}_i$), \ie, the one-step-ahead location predicted by Bearing-UAV.
Then, we obtain current UVP $\bm{I}_i^U$ using $\bm{r}_i$ from the UAV-view satellite image, while the corresponding RSB is simultaneously retrieved in the form of onboard RST features $\bm{B}_i=\{\bm{t}_{i,j}\}_{j=1}^4$ according to index $\bm{n}_i$.
We then perform cross-view regression by:
\begin{equation}
    (\hat{\bm{p}}_i, \hat{\bm{h}}_i) := \mathcal{F}_\mathrm{Bearing-UAV} \left(\bm{I}_i^U, \bm{B}_i, \bm{\mathcal{C}}\right) \text{.}
\end{equation}

Given the current waypoint, we compute the azimuth ${a}_i$ from the UAV to the next waypoint for the next step, adjust the UAV's heading to align with azimuth ${a}_i$, and update $(\bm{r}_{i+1}, \bm{n}_{i+1})$ accordingly and proceed to the next iteration.

Accurate heading alignment is crucial for purely vision-based navigation because horizontal rotational drift often occurs during long-range flights, and without reliable heading estimation, the UAV heading is difficult to align with the reference azimuth, leading to navigation drift.

\section{Experiments}
\label{sec:4_experiment1}

\subsection{Dataset}
To evaluate purely vision-based UAV localization and navigation under unaligned U-S cross-view settings, we construct a new dataset, \textbf{\dataset}, as shown in~\cref{table:dset}. 

We collect samples from Google Earth under two modes. 
In Google Earth 2D mode (satellite-view mode), we first download four contiguous satellite images from four cities and crop them into RSTs. 
Each image (4096$\times$4096 pixels at 0.25 m/px) is partitioned into $16\times16$ RSTs. 
Any 2$\times$2 block of adjacent RSTs forms an RSB, yielding 15$\times$15 indexed RSBs. 
In Google Earth 3D mode (UAV-view mode), we directly sample UVPs over the same area. 
For each RSB, we sample 100 random camera positions and yaw angles via viewpoint roaming, resulting in 90k UVPs. 
Each UVP is paired with a JSON file containing geographic information. We also collect 90k satellite-view patches as an ideal reference.
See \supp\ref{sec:dataconstruction} and \cref{fig:city1234} therein for more details on satellite images, dataset construction, and data licensing.

\begin{table}[!t]
\centering
\setlength{\tabcolsep}{2.3pt}  
\caption{Comparison between \dataset and other geo-localization datasets. 
Existing datasets rarely consider unaligned scenarios. 
Ours focuses on arbitrary rotations (with varying IoUs and challenging misalignment) between UVPs and RSTs, and provides heading-aware annotations, forming a more comprehensive testbed for UAV localization and navigation.
SUES = SUES-200, Dense = DenseUAV, Contiguous\textsuperscript{*}: RSTs form a contiguous map.}
\label{table:dset}
\small  
\begin{tabular}{@{}lrrrrr@{}}  
    \toprule
    \textbf{Datasets} & \textbf{Uni-1652} & \textbf{SUES} & \textbf{Dense} & \textbf{GTA-UAV} & \textbf{Ours} \\
    \midrule
    \#UAV              & 37{,}854 & 24{,}210 & 18{,}198 & 33{,}763 & 90k$\times$2 \\
    \#Satellite        & 1{,}652  & 200      & 9{,}096  & 9{,}096  & 1{,}024 \\
    Scene              & Building & Campus   & Campus   & City     & Multi-city \\
    GeoTag             & Fine     & Coarse   & Fine     & Synthetic& Fine \\
    Contiguous\textsuperscript{*} & \ding{55} & \ding{55} & \ding{51} & \ding{51} & \ding{51} \\
    Unaligned          & \ding{55} & \ding{55} & \ding{55} & \ding{51} & \ding{51} \\
    Heading            & \ding{55} & \ding{55} & \ding{55} & \ding{55} & \ding{51} \\
    \bottomrule
\end{tabular}
\end{table}

To the best of our knowledge, there is no public U-S cross-view, multi-city dataset with contiguous satellite tiles and abundant unaligned UAV views, that is specifically designed for purely vision-based UAV localization and navigation, as shown in ~\cref{table:dset}. 
In contrast, \dataset offers U-S cross-view discrete samples for retrieval/matching-based localization, heading annotations for orientation evaluation, and a navigation benchmark. 
Based on the multi-city maps, we design eight curved navigation routes with multiple waypoints, and leverage contiguous cross-city overhead imagery together with Google Earth 3D mode to provide a realistic platform for evaluating purely vision-based UAV navigation.

\begin{table*}[!t]
\centering
\setlength{\tabcolsep}{2.7pt}
\caption{The geo-localization and navigation performance in satellite views and UAV views (U-S cross-view). WA: weather augmentation.}
\label{tab:table_locnav}
\small
\begin{tabular}{l *{16}{c}}
    \toprule
    & \multicolumn{10}{c}{\textbf{Geo-Localization}}
    & \multicolumn{6}{c}{\textbf{Navigation}} \\
    \cmidrule(lr){2-11} \cmidrule(lr){12-17}

    \multirow[t]{2}{*}{\textbf{Method}}
    & \multicolumn{2}{c}{\textbf{Recall@1}\textsuperscript{\scriptsize$\uparrow$}}
    & \multicolumn{2}{c}{\textbf{LSR@15}\textsuperscript{\scriptsize$\uparrow$}}
    & \multicolumn{2}{c}{\textbf{HSR@15}\textsuperscript{\scriptsize$\uparrow$}}
    & \multicolumn{2}{c}{\textbf{MLE}\textsuperscript{\scriptsize$\downarrow$}}
    & \multicolumn{2}{c}{\textbf{MHE}\textsuperscript{\scriptsize$\downarrow$}}
    & \multicolumn{2}{c}{\textbf{SR@20}\textsuperscript{\scriptsize$\uparrow$}}
    & \multicolumn{2}{c}{\textbf{SPL}\textsuperscript{\scriptsize$\uparrow$}}
    & \multicolumn{2}{c}{\textbf{NE}\textsuperscript{\scriptsize$\downarrow$}} \\
    \cmidrule(lr){2-3}\cmidrule(lr){4-5}\cmidrule(lr){6-7}\cmidrule(lr){8-9}\cmidrule(lr){10-11}
    \cmidrule(lr){12-13}\cmidrule(lr){14-15}\cmidrule(lr){16-17}
    & \multicolumn{1}{c}{Sat.} & \multicolumn{1}{c}{UAV}
    & \multicolumn{1}{c}{Sat.} & \multicolumn{1}{c}{UAV}
    & \multicolumn{1}{c}{Sat.} & \multicolumn{1}{c}{UAV}
    & \multicolumn{1}{c}{Sat.} & \multicolumn{1}{c}{UAV}
    & \multicolumn{1}{c}{Sat.} & \multicolumn{1}{c}{UAV}
    & \multicolumn{1}{c}{Sat.} & \multicolumn{1}{c}{UAV}
    & \multicolumn{1}{c}{Sat.} & \multicolumn{1}{c}{UAV}
    & \multicolumn{1}{c}{Sat.} & \multicolumn{1}{c}{UAV} \\
    \midrule

    University-1652 \cite{zheng2020university1652}
        & 63.29 & 60.20
        & 16.31 & 15.11
        &   --  &   --
        & 31.78 & 33.15
        &   --  &   --
        &  0.00 &  0.00
        &  0.00 &  0.00
        &  546.33 &  602.96 \\

    SUES-200 \cite{zhu2023sues200}
        & 71.96 & 66.60
        & 16.78 & 15.76
        &   --  &   --
        & 29.09 & 30.83
        &   --  &   --
        &  0.00 &  0.00
        &  0.00 &  0.00
        &  439.50 & 618.85 \\

    DenseUAV \cite{dai2024denseuav2}
        & 80.04 & 73.43
        & 17.23 & 16.54
        &   --  &   --
        & 26.82 & 28.79
        &   --  &   --
        & 0.00 & 0.00
        & 0.00 & 0.00
        & 627.64 & 651.93 \\

    GTA-UAV \cite{ji2025game4loc}
        & 76.99 & 70.71
        & 31.33 & 27.96
        &   --  &   --
        & 25.24 & 28.43
        &   --  &   --
        & 0.00 & 0.00
        & 0.00 & 0.00
        & 666.51 & 661.91 \\
    \midrule

    Ours VGG-16
        & 90.76 & 83.17
        & 98.33 & 89.36
        & 98.12 & 77.21  
        &  5.66 &  8.61
        &  \textbf{4.15} & 12.90
        & \textbf{62.50} & \textbf{50.00}
        & \textbf{45.18} & \textbf{29.82}
        & \textbf{25.61} & 275.61  \\

    Ours VGG-16 + WA
        & \textbf{91.07} & \textbf{86.52}
        & \textbf{98.63} & \textbf{92.88}
        & \textbf{98.21} & \textbf{86.02}  
        &  \textbf{5.40} &  \textbf{7.48}
        &  4.21 &  \textbf{9.63}
        & 37.50 & 25.00
        & 27.57 & 16.71
        & 325.00 & \textbf{248.77} \\
        
    \bottomrule
\end{tabular}
\label{table:locnav}
\end{table*}

\subsection{Implementation}
\label{sec:4_experiment2}

\paragraph{Network Configuration}
We adopt VGG-16~\cite{Simonyan2015VeryDeepCNN} pretrained on ImageNet~\cite{deng2009imagenet} as the visual backbone for all experiments unless otherwise specified. 
We set the number of clusters in the GLUF to $K=4$ and base
feature dimension to $D=256$. For RCE, we adopt a layer configuration
$[d_1,\dots,d_L] = [2,64,256,KD]$, and the dual regressor branches use an MLP
with dimensions $[2050,1024,256,64,2]$.

We set $\boldsymbol{\mathcal{C}} = \{(-1,1), (-1,-1), (1,1), (1,-1)\}$ as the relative coordinates of four RSTs in each RSB, such that, given the RSB index,
the absolute geo-localization can be recovered deterministically, while the network
only regresses a bounded, dimensionless target. This parameterization stabilizes
optimization and circumvents the difficulties of directly regressing
high-precision latitude/longitude values.

\paragraph{Training Setup}
The dataset is split 7:2:1 for training, validation, and test. 
We use Adam (lr=5×10$^{-5}$, batch size 16) for 100 epochs with Smooth L1 loss:
$\mathcal{L}_{sum}=0.8\mathcal{L}_p+0.2\mathcal{L}_h$, without weight decay. 
A ReduceLROnPlateau scheduler halves the learning rate upon a validation plateau, 
and the best model is selected by validation loss. 
Training and evaluation are conducted on an NVIDIA H100 GPU, 
while Bearing-Naver runs on a laptop with an RTX 4000 GPU.

\subsection{Experimental Results}
\label{sec:4_experiment3}

\subsubsection{Evaluation Protocol and Setup}
\begin{figure}[!t]
\centering%
    \includegraphics[width=\linewidth]{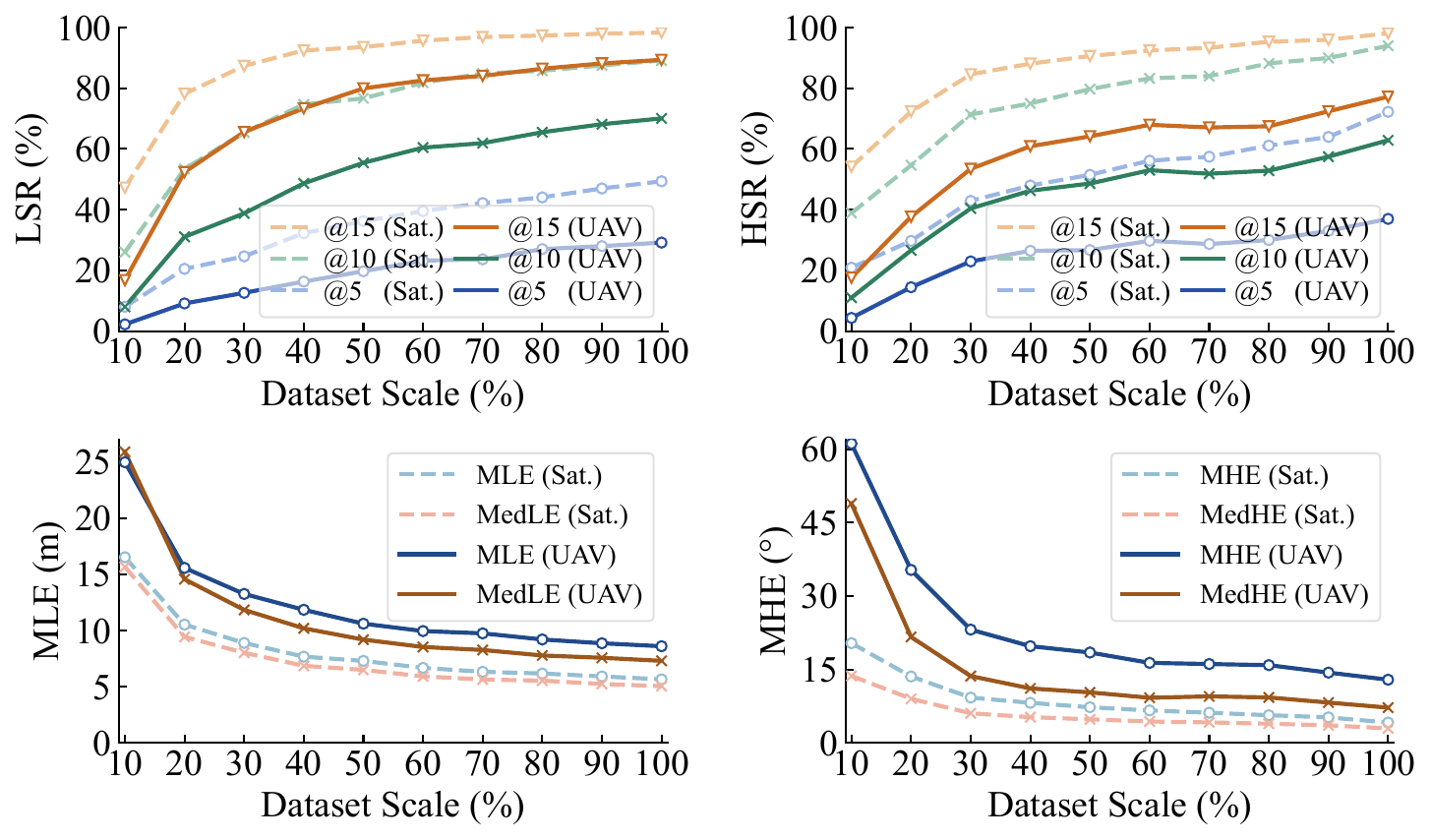}
\caption{Localization and heading performance both consistently improve with increasing dataset scale and gradually begin to saturate once the dataset scale exceeds 60\% of \dataset.}%
\label{fig:dsetscale}%
\end{figure}

For localization/heading estimation, we report Recall@K, LSR/HSR, and MLE/MHE. For navigation, we report SR, SPL, and NE. We also report model size, inference time, and GFLOPs. See \supp\ref{sec:acronyms} for metric definitions.

Experiments include comparisons with existing CVGL approaches, 
backbone replacements, analysis of dataset scale and city diversity, 
evaluation of weather augmentation, and long-range navigation with multiple waypoints. 
We also conduct parallel satellite-view localization and navigation experiments as an ideal-case reference benchmark.

\subsubsection{Localization and Heading Performance}
We summarize the Geo-Localization results in the first five columns of~\cref{tab:table_locnav}.
The localization performance of four representative CVGL baselines improves from University-1652 to GTA-UAV, but our method consistently achieves the best results.
All baselines lack heading estimation ability and exhibit localization errors around 30\,m, far larger than our regression error of 8.6\,m. This is mainly because they treat the retrieved or matched tile center as the final position, which struggles under cross-view misalignment and varying IoUs. 
Our method improves SR@1 by $\sim$10\% and LSR@15 by $\sim$60\%, demonstrating a stronger ability to identify the RST closest to the UVP and accurately localize the UVP, reflecting the benefit of the regression paradigm.

Notably, weather-augmented training improves performance, reducing MLE by $1.1\,$m and MHE by $3.3^\circ$, see Sec.~\ref{sec:expweather} for details.
We also extensively evaluate additional backbones (ResNet, ViT, MobileNet) in \supp\ref{sec:backbone}, and analyze the localization/heading error distribution and their causal factors at three granularity levels in \supp\ref{sec:locerrstat}.

\subsubsection{Effect of Dataset Scale on Bearing-UAV}

\dataset provides 100 samples per RSB, resulting in 90k UAV-view patches. 
To study the impact of dataset size to our model, we vary the sampling rate and construct ten datasets 
of increasing scale for both satellite and UAV views. As shown in ~\cref{fig:dsetscale}, 
localization and heading performance improve consistently with more training data. 
When the dataset size reaches 54k (60\% of the full dataset), the gains begin to saturate: 
in the UAV-view setting, MLE falls below 10\,m and MHE below 17$^\circ$, 
while in the satellite-view setting, the model achieves below 7$\,\mathrm{m}$ MLE and 7$^\circ$ MHE.
Correspondingly, LSR and HSR also show a gradual convergence trend.
For example, at 15\,m/$^\circ$ success radius, they exceed 80\% and 65\% respectively once the dataset scale surpasses 60\%, with the satellite-view curve exhibiting a higher and smoother trend.

\begin{table}[!t]
\centering
\setlength{\tabcolsep}{2.5pt}  
\caption{Effect of multi-city diversity on model generalization performance. 
We design four distinct city combinations for training.}
\label{tab:table_city1234}
\begin{tabular}{@{}lcccccccc@{}}
    \toprule
    \multirow{2}{*}{\textbf{Cities}}
    & \multicolumn{2}{c}{\textbf{Recall@1}\textsuperscript{\scriptsize$\uparrow$}}
    & \multicolumn{2}{c}{\textbf{LSR@15}\textsuperscript{\scriptsize$\uparrow$}}
    & \multicolumn{2}{c}{\textbf{MLE}\textsuperscript{\scriptsize$\downarrow$}}
    & \multicolumn{2}{c}{\textbf{MHE}\textsuperscript{\scriptsize$\downarrow$}} \\
    \cmidrule(lr){2-3}\cmidrule(lr){4-5}\cmidrule(lr){6-7}\cmidrule(lr){8-9}
    & \multicolumn{1}{c}{Sat.} & \multicolumn{1}{c}{UAV}
    & \multicolumn{1}{c}{Sat.} & \multicolumn{1}{c}{UAV}
    & \multicolumn{1}{c}{Sat.} & \multicolumn{1}{c}{UAV}
    & \multicolumn{1}{c}{Sat.} & \multicolumn{1}{c}{UAV} \\
    \midrule

    A
        & 89.43	& 79.92	& 97.20	& 80.36	& 6.14	& 10.29	& 4.92	& 12.07 \\
    B
        & 88.76	& 83.39	& 96.36	& 88.14	& 6.44	& 8.97	& 4.28	& \textbf{10.27} \\
    C
        & 90.40	& \textbf{85.43}	& 98.22	& 88.01	& 5.54	& 8.67	& \textbf{4.00}	& 17.15 \\
    D
        & 88.18	& 78.90	& 96.18	& 82.72	& 6.62	& 10.20	& 4.90	& 15.71 \\
    \midrule
    AD
        & 89.58	& 80.54	& 97.96	& 83.14	& 6.03	& 10.00	& 4.97	& 14.65 \\
    BC
        & 90.22	& 85.23	& \textbf{98.47}	& \textbf{91.02}	& \textbf{5.46}	& \textbf{8.21}	& 4.70	& 13.08 \\
    \midrule
    ABD
        & 89.93	& 81.57	& 97.97	& 86.44	& 5.91	& 9.20	& 5.01	& 12.82 \\
    BCD
        & 90.24	& 84.44	& 98.36	& 90.62	& 5.70	& 8.32	& 4.56	& 13.86 \\
    \midrule
    ABCD
        & \textbf{90.76}	& 83.17	& 98.33	& 89.36	& 5.66	& 8.61	& 4.15	& 12.90 \\
    \bottomrule
\end{tabular}
\end{table}

\subsubsection{Effect of City Combinations on Bearing-UAV}

Unlike simply scaling the dataset size, this experiment focuses on how well the model adapts to diverse urban terrains and layouts. See \supp\cref{fig:city1234} for satellite imagery details.

We train our model on datasets constructed from different combinations of these cities. 
The results are summarized in~\cref{tab:table_city1234} and \supp\cref{fig:multicitycurve}. 
From top to bottom, the four groups correspond to datasets of 1, 2, 3, and 4 cities respectively, with increasing diversity.
In the single-city group, the satellite-view setting is relatively stable across cities, since there is no spatial visual discrepancy, only misalignment. 
In contrast, the UAV-view setting orientation performance varies significantly. City~C shows the largest errors, followed by D, while B performs best. This is mainly because tall buildings in City~C induce strong cross-view appearance changes, and the river region provides limited texture. City~D combines mountainous areas with many similarly structured small buildings, resulting in limited visual distinctiveness and thus larger localization and heading errors. City~B offers rich, distinctive building patterns, while City~A, although vegetation-dominant, still contains more diverse textures than City~D, yielding slightly better results.

These trends are consistent in the two-city combinations: the BC pair outperforms the AD pair, supporting the above observations. For three-city combinations, mixing heterogeneous cities reduces variance, and the performance gap between different triplets becomes smaller, similar to the results observed between City~A and City~C.
Further details are discussed in \supp\ref{sec:dsetscale}.

Most notably, as the number of cities increases from 1 to 4, the overall performance does not degrade despite larger inter-city variations and increased scene complexity; the averaged metrics even show a slight improvement. This indicates that our model generalizes well across diverse multi-city environments and benefits from richer geographic diversity.

\begin{table}[!t]
\centering
\setlength{\tabcolsep}{1.5pt} 
\caption{Weather robustness. Rows 1–6: augmented model under six conditions; Baseline: no augmentation under normal weather.
}
\label{table:weather}

\begin{tabular}{@{}l *{8}{c}@{}}
    \toprule
    \multirow{2}{*}{\textbf{Weather}}   
    & \multicolumn{2}{c}{\textbf{LSR@15}\textsuperscript{\scriptsize$\uparrow$}} 
    & \multicolumn{2}{c}{\textbf{HSR@15}\textsuperscript{\scriptsize$\uparrow$}}
    & \multicolumn{2}{c}{\textbf{MLE}\textsuperscript{\scriptsize$\downarrow$}}
    & \multicolumn{2}{c}{\textbf{MHE}\textsuperscript{\scriptsize$\downarrow$}}\\
    \cmidrule(lr){2-3} \cmidrule(lr){4-5} \cmidrule(lr){6-7} \cmidrule(lr){8-9}
      & \textbf{Sat.} & \textbf{UAV}
      & \textbf{Sat.} & \textbf{UAV}
      & \textbf{Sat.} & \textbf{UAV}
      & \textbf{Sat.} & \textbf{UAV} \\
    \midrule
    
    Rain       & 95.63 & 89.75 & 93.65 & 79.06 & 6.70 &  8.41 &  6.15 & 12.04 \\
    Snow       & 94.80 & 88.79 & 94.33 & 79.28 & 6.95 &  8.64 &  6.04 & 12.04 \\
    Fog        & 95.61 & 89.02 & 93.86 & 78.72 & 6.68 &  8.55 &  6.36 & 12.56 \\
    Bright     & 98.02 & 92.25 & 97.59 & 83.71 & 5.72 &  7.69 &  4.68 & 10.31 \\
    Mixed      & 96.34 & 90.33 & 95.44 & 81.29 & 6.35 &  8.19 &  5.59 & 11.24 \\
    Normal     & \textbf{98.63} & \textbf{92.88} & \textbf{98.21} & \textbf{86.02} & \textbf{ 5.40} & \textbf{ 7.48} & 4.21 & \textbf{ 9.63} \\
    \midrule
    Baseline   & 98.33 & 89.36 & 98.12 & 77.21 &  5.66 &  8.61 & \textbf{ 4.15} & 12.90 \\ 

    \bottomrule
\end{tabular}
\end{table}

\subsubsection{Weather Augmentation Test}
\label{sec:expweather}
To assess weather effects, we augment \dataset with illumination, fog, rain, and snow (20\% each), as shown in \supp\cref{fig:supweatheraug}, to train a weather-augmented model, which we evaluate on six weather conditions and compare against the non-augmented baseline in~\cref{table:weather} and \supp\cref{fig:supweathercurve}.

Across all four metrics, especially in the U-S cross-view, the augmented model consistently outperforms the non-augmented baseline across various weather conditions in both localization and heading estimation. 
This indicates that expanding the training distribution via weather augmentation improves generalization. More importantly, it suggests that the model learns a shared, weather-robust representation, benefiting from diverse weather exposure rather than overfitting to a single appearance pattern.
It is also worth noting that illumination augmentation yields the most significant performance gain. 
Given the large brightness discrepancy between UAV and satellite views, such augmentation effectively reduces the cross-view illumination gap for cross-view localization.
See \supp\ref{sec:weatheraug} for further discussion.

\subsubsection{Bearing-Naver Navigation Test}

We design two routes per city, each with a length between 500\,m and 1200\,m and more than ten waypoints covering diverse scene types. Assume the UAV uses a step size of 25 m, and the threshold radius for reaching a waypoint is 20 m. 
The test results are reported in~\cref{table:locnav}. 
Most baseline methods fail to complete the full route, their localization errors cause drift or dithering, especially in feature-sparse or visually confusing regions. In contrast, our method achieves high-precision localization and completes nearly half of the challenging, tortuous routes.
Compared with Ours VGG-16, the reduced SR and SPL of the weather-augmented model are partly due to several highly deviated trajectories.
Yet the NE drops from 275\,m to 248\,m, indicating that the UAV reaches the target more closely in more cases and thus has an improved ability to reach the goal.

\begin{figure}[!t]
\centering%
    \includegraphics[width=\linewidth]{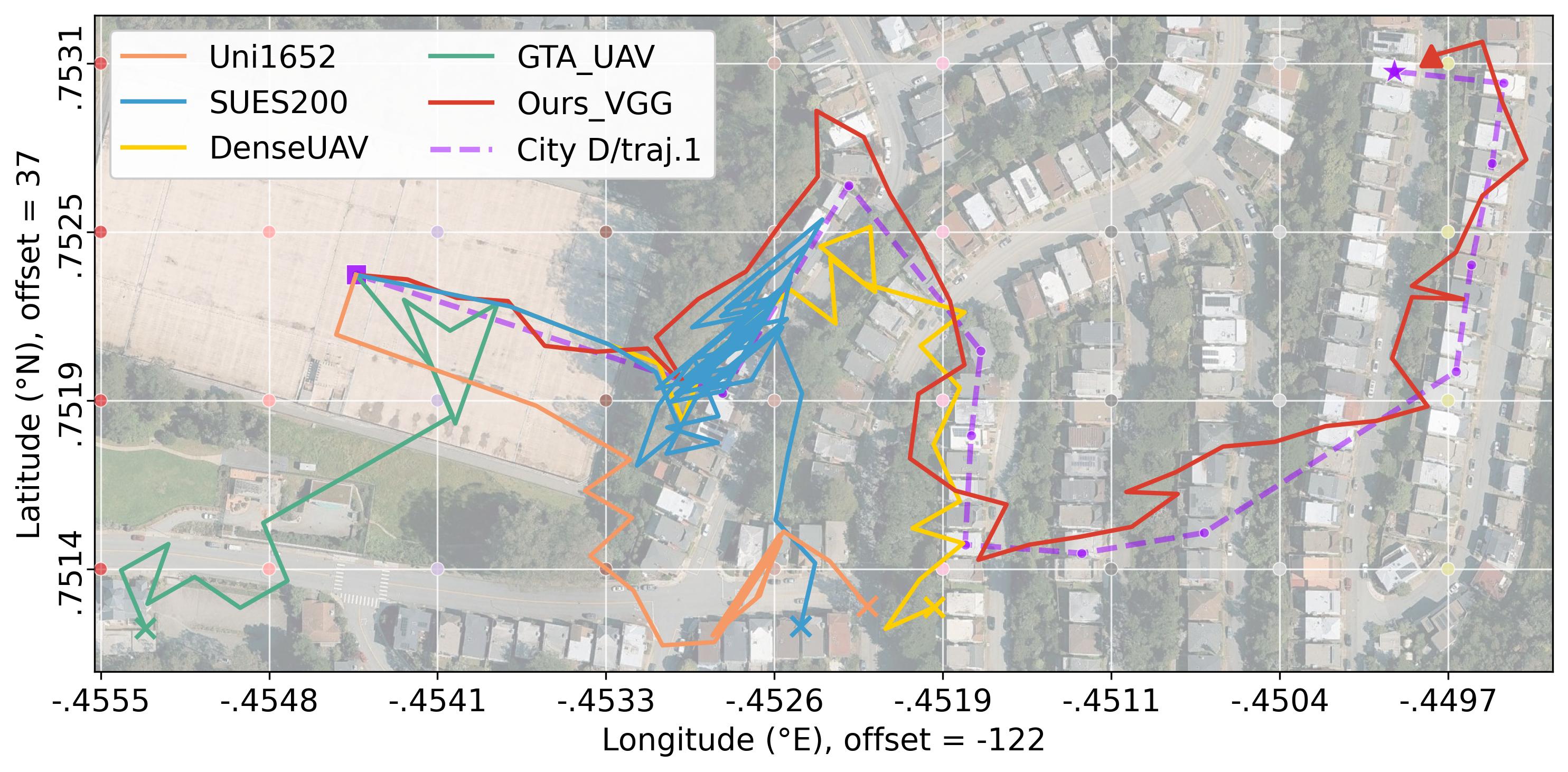}
\caption{Navigation performance comparison. The purple dashed line here denotes the predefined navigation trajectory \#1 in City D. $\scriptstyle\blacksquare$ start, \scalebox{0.8}{$\bigstar$} end, × failure, $\blacktriangle$ success. 
The corresponding UAV-view frames across the 45 navigation steps are shown in \supp\cref{fig:ourvggd50frames3d}.}%
\label{fig:navtraj}
\end{figure}

In~\cref{fig:navtraj}, trajectory \#1 in City D is about 720\,m long and contains 13 waypoints. It starts from the square marker in a feature-sparse open area, passes over dozens of similar-looking buildings and green spaces along a tortuous path, and ends at the star marker over a typical white rooftop. 
On this challenging route, only our method successfully reaches the final waypoint within 45 steps. DenseUAV completes roughly half of the path, SUES-200 hovers over a row of rooftops near the beginning and then drifts away, while University-1652 and GTA-UAV deviate from the correct heading almost immediately after takeoff. During navigation, the UAV views and the satellite views contain many unaligned U-S scenes with rapidly changing IoUs, compounded by cross-view parallax and feature-sparse regions, so methods with lower localization accuracy are much more likely to fail under this setting.
The remaining seven trajectories and analysis are shown in \supp\ref{sec:navtest}.

We report on-disk model size, single-step inference time, and GFLOPs for the five models in~\cref{table:perform}. 
\method is lightweight, achieves near real-time performance, and scales in constant time to larger maps and longer paths.

\begin{table}[!t]
\centering
\setlength{\tabcolsep}{1.8pt}  
\renewcommand{\arraystretch}{1.0}   
\caption{
Model efficiency. MS: model size; IT: inference time. GFLOPs are computed at 256$\times$256 or 384$\times$384 as in prior work.
}
\label{table:perform}
\small
\begin{tabular}{@{}lrrrrr@{}}  
    \toprule
    \textbf{Metric} &
    \textbf{Uni-1652} &
    \textbf{SUES} &
    \textbf{Dense} &
    \textbf{GTA-UAV} &
    \textbf{Ours} \\
    \midrule

    MS\textsuperscript{\scriptsize$\downarrow$} (MB)
    & 102 & 658 & 86 & 326 & \textbf{66} \\

    IT\textsuperscript{\scriptsize$\downarrow$} (ms)
    & \textbf{10.1} & 1907.5 & 593.6 & 246.5 & 133.5 \\

    GFLOPs\textsuperscript{\scriptsize$\downarrow$}
    & 16.4{\footnotesize $@$}256 & 98.7{\footnotesize $@$}384 & \textbf{8.5}{\footnotesize $@$}256 & 20.1{\footnotesize $@$}384 & 10.5{\footnotesize $@$}256 \\
    
    \bottomrule
\end{tabular}
\end{table}

\subsection{Ablation Study}

In this section, we conduct ablation studies on the key submodules of our model: GLUF, RCE, PSG, and CA. 
As summarized in~\cref{tab:table_ablation}, we report five metrics on UAV-view setting, with the satellite-view setting reported in \supp\ref{sec:ablation}.
Removing GLUF causes a significant drop in performance, 
indicating that clustering and recombining feature maps helps extract more robust local and global structures, 
which is crucial for localization and heading estimation under cross-view misalignment.
RCE has a clear impact on orientation: it reduces MHE by approximately $2^\circ$ and improves HSR@15 by about 6.5\%,
showing that injecting positional embeddings into RSTs benefits heading regression.
PSG and CA each contribute an additional 1--2 percentage points to localization and heading success rates,
demonstrating that they further strengthen feature alignment.

\begin{table}[!t]
\centering
\setlength{\tabcolsep}{1.5pt}         
\renewcommand{\arraystretch}{1.0}   
\caption{Ablation study of the \method on the UAV-view data.}
\label{tab:table_ablation}
\begin{tabular}{@{}llllccccc@{}}
    \toprule
    \textbf{GLUF} & \textbf{RCE} & \textbf{PSG} & \textbf{CA}
    & \textbf{R@1}\textsuperscript{\scriptsize$\uparrow$}
    & \textbf{LSR}\textsuperscript{\scriptsize$\uparrow$}
    & \textbf{HSR}\textsuperscript{\scriptsize$\uparrow$}
    & \textbf{MLE}\textsuperscript{\scriptsize$\downarrow$}
    & \textbf{MHE}\textsuperscript{\scriptsize$\downarrow$} \\
    \midrule

    \ding{55} & \ding{51} & \ding{51} & \ding{51} & 67.19 & 57.29 & 61.10 & 14.90 & 21.62 \\
    \ding{51} & \ding{55} & \ding{51} & \ding{51} & 82.66 & 87.77 & 70.80 &  8.93 & 14.99 \\
    \ding{51} & \ding{51} & \ding{55} & \ding{51} & \textbf{83.97} & 88.68 & 75.28 &  8.67 & 13.44 \\
    \ding{51} & \ding{51} & \ding{51} & \ding{55} & 83.70 & 89.06 & 76.68 &  \textbf{8.59} & 13.20 \\
    \ding{51} & \ding{51} & \ding{51} & \ding{51} & 83.17 & \textbf{89.36} & \textbf{77.21} &  8.61 & \textbf{12.90} \\
    \bottomrule
\end{tabular}
\end{table}

\section{Conclusion and Future Work}
\label{sec:conclusion}

In this paper, we go beyond standard CVGL by jointly estimating accurate geo-localization and reliable heading for GNSS-denied UAV navigation. Our goal is to jointly infer position and heading under cross-view, non-aligned, and arbitrarily rotated U-S image pairs, using only vision, without any auxiliary sensors or additional geometric reasoning. To this end, we propose a single-stage regression network that captures both global and local structural cues and explicitly encodes relative spatial relationships, enabling robustness to viewpoint-induced parallax, misalignment, varying IoUs, and sparse visual features. Besides, a benchmark cross-view, multi-city dataset and comprehensive evaluation metrics are constructed. Extensive experiments have shown encouraging results that the regression framework can reliably perform cross-view localization and heading estimation in complex scenarios. 
Limitations are discussed in \supp\ref{sec:suppl_lim}.
\section{Acknowledgements}
\label{sec:acknowledgements}

This work is supported by the Hangzhou Joint Fund of the Zhejiang Provincial Natural Science Foundation of China (Grant No. LHZSD24F020001), and Zhejiang Province High-Level Talents Special Support Program ``Leading Talent of Technological Innovation of Ten-Thousands Talents Program" (No.2022R52046).

{
    \small
    \bibliographystyle{ieeenat_fullname}
    \clearpage  
    \bibliography{references/cvphr}
}


\clearpage 
\appendix %
\setcounter{section}{0} 
\renewcommand{\thesection}{A.\arabic{section}} 
\renewcommand{\thesubsection}{A.\arabic{section}.\arabic{subsection}} 

\clearpage
\setcounter{page}{1}
\maketitlesupplementary


\section{List of Acronyms}
\label{sec:acronyms}
For clarity, the main acronyms used in this paper are grouped into four categories and summarized in~\cref{tab:table_acronym}.
\begin{table}[H]
\centering
\setlength{\tabcolsep}{4pt}
\renewcommand{\arraystretch}{1.05}
\caption{List of main acronyms used in this paper.}
\label{tab:table_acronym}
\begin{tabular}{llr}
    \toprule
    \textbf{Type} & \textbf{Acronym} & \textbf{Description} \\
    \midrule
    \multirow{4}{*}{Concept}
    & UAV & Unmanned Aerial Vehicle \\
    & GNSS & Global Navigation Satellite System \\
    & CVGL & Cross-View Geo-Localization \\
    & M2T & Match-to-Tile \\
    \midrule
    \multirow{5}{*}{Data}
    & UVP & UAV-View Patch \\
    & RST & Remote-Sensing Tile \\
    & RSB & Remote-Sensing Block \\
    & U-S & UAV-Satellite \\
    & G-S & Ground-Satellite \\
    \midrule
    \multirow{4}{*}{Module}
    & GLUF & Global-Local Unity Feature \\
    & RCE & Relative Coordinate Encoder \\
    & PSG & Patch Similarity-Guided \\
    & CA & Cross-Attention \\
    \midrule
    \multirow{10}{*}{Metric}
    & Recall@1 & Recall at 1 \\
    & LSR & Localization Success Rate \\
    & HSR & Heading Success Rate \\
    & MLE & Mean Localization Error \\
    & MHE & Mean Heading Error \\
    & MedLE & Median Localization Error \\
    & MedHE & Median Heading Error \\
    & SR@20 & Success Rate at 20 m \\
    & SPL & Success Weighted by Path Length \\
    & NE & Navigation Error \\
    \bottomrule 
\end{tabular}
\end{table}

\section{Localization and Heading Performance with Different Backbones}
\label{sec:backbone} 
We evaluate \method with four different backbone networks, as shown in~\cref{tab:table_locbone}.
Here, we define Recall@1 as the accuracy of retrieving the RST whose location is closest to the UVP from the four adjacent RSTs.
Each metric is reported in two forms, “Sat.” and “UAV”, denoting performance under the satellite view (Sat.) and the UAV view (U-S cross-view), respectively. 
The former serves as a near-ideal reference benchmark, while the latter corresponds to our target cross-view localization task.

\begin{table*}[!t]
\centering
\setlength{\tabcolsep}{4pt}
\caption{Comparison of localization and heading performance with four different backbones in both satellite views and UAV views (U-S cross-view). \method with the VGG-16 backbone achieves the best overall results. Mobile-V3S = MobileNet-V3-Small.}
\label{tab:table_locbone}
\begin{tabular}{l *{14}{r}}
    \toprule
    \multirow[t]{2}{*}{\textbf{Method}}
    & \multicolumn{2}{c}{\textbf{Recall@1}\textsuperscript{\scriptsize$\uparrow$}}
    & \multicolumn{2}{c}{\textbf{LSR@15}\textsuperscript{\scriptsize$\uparrow$}}
    & \multicolumn{2}{c}{\textbf{HSR@15}\textsuperscript{\scriptsize$\uparrow$}}
    & \multicolumn{2}{c}{\textbf{MLE}\textsuperscript{\scriptsize$\downarrow$}}
    & \multicolumn{2}{c}{\textbf{MedLE}\textsuperscript{\scriptsize$\downarrow$}}
    & \multicolumn{2}{c}{\textbf{MHE}\textsuperscript{\scriptsize$\downarrow$}}
    & \multicolumn{2}{c}{\textbf{MedHE}\textsuperscript{\scriptsize$\downarrow$}} \\
    \cmidrule(lr){2-3}\cmidrule(lr){4-5}\cmidrule(lr){6-7}\cmidrule(lr){8-9}\cmidrule(lr){10-11}\cmidrule(lr){12-13}\cmidrule(lr){14-15}
    & \multicolumn{1}{c}{Sat.} & \multicolumn{1}{c}{UAV}
    & \multicolumn{1}{c}{Sat.} & \multicolumn{1}{c}{UAV}
    & \multicolumn{1}{c}{Sat.} & \multicolumn{1}{c}{UAV}
    & \multicolumn{1}{c}{Sat.} & \multicolumn{1}{c}{UAV}
    & \multicolumn{1}{c}{Sat.} & \multicolumn{1}{c}{UAV}
    & \multicolumn{1}{c}{Sat.} & \multicolumn{1}{c}{UAV}
    & \multicolumn{1}{c}{Sat.} & \multicolumn{1}{c}{UAV} \\
    \midrule

    Ours ResNet18
        & 83.88 & 75.34
        & 88.38 & 71.83
        & 73.20 & 46.83
        &  8.83 & 12.09
        &  7.89 & 10.56
        & 14.59 & 26.52
        &  8.54 & 16.16 \\

    Ours ViT-Small
        & 86.38 & 81.39
        & 93.35 & 85.47
        & 77.58 & 51.57
        &  7.53 &  9.46
        &  6.73 &  8.02
        & 12.02 & 24.89
        &  7.69 & 14.45  \\

    Ours Mobile-V3S
        & 86.41 & 79.76
        & 93.52 & 81.20
        & 83.25 & 64.94
        &  7.49 & 10.34
        &  6.73 &  8.72
        & 10.57 & 19.53
        &  6.28 & 10.14   \\

    Ours VGG-16
        & \textbf{90.76} & \textbf{83.17}
        & \textbf{98.33} & \textbf{89.36}
        & \textbf{98.12} & \textbf{77.21}
        &  \textbf{5.66} &  \textbf{8.61}
        &  \textbf{5.05} &  \textbf{7.30}
        &  \textbf{4.15} & \textbf{12.90}
        &  \textbf{2.96} & \textbf{7.20} \\

    \bottomrule
\end{tabular}
\end{table*}

By comparing the four rows in~\cref{tab:table_locbone}, we observe that using ResNet18 or ViT-Small as the backbone leads to noticeably worse performance, MobileNet-V3-Small is slightly better than ViT-Small, whereas VGG-16 achieves a clear advantage on all metrics.

With comparable model sizes, ViT-Small and ResNet18 are better at capturing high-level global semantics, which is advantageous when features are well aligned, but less suitable under misaligned cross-view conditions where fine local cues are critical.
In our localization and navigation setting, the feature maps produced by the backbone are further processed by the GLUF module to extract local features for misalignment-robust retrieval and localization.
The reduced low-level structural information in these deeper, more global backbones limits their localization accuracy and thus constrains the overall performance.
Moreover, ViT-Small is known to be data-hungry; given the moderate scale of our dataset and the domain gap between ImageNet-style pre-training and overhead/aerial imagery, its potential is not fully realized. 

In contrast, simpler CNN backbones such as VGG-16 (and MobileNet-V3S) preserve richer fine-grained spatial details, supplying the GLUF module with stronger local structural cues that are essential for robust UAV-satellite feature matching, especially under misaligned and varying-IoU conditions.
This suggests that classical convolutional backbones can still offer distinct advantages in cross-view localization-orientation scenarios.

Another noteworthy observation is that MedLE and MedHE are consistently lower than the corresponding MLE and MHE, and this gap is more pronounced in the cross-view setting, where the median localization error is 1.3\,m lower than the mean and the median heading error is 5.7$^\circ$ lower than the mean.
This phenomenon is closely related to the long-tailed distribution of localization and heading errors, especially for heading, where a small portion of unseen samples with large errors substantially inflates the mean, and the trend is as illustrated in~\cref{fig:locerrstat}.

\section{Cross-view Multi-city Dataset Design}
\label{sec:dataset2}
All data of our \dataset were collected from Google Earth.
Our use of Google Earth maps strictly follows its terms for non-commercial academic research, consistent with prior datasets such as DOTA (CVPR 2018) and OmniCity (CVPR 2023).
\dataset is publicly available at \url{https://huggingface.co/datasets/HaoyZhou/bearinguav/tree/main}.

\subsection{Dataset Construction Process}
\label{sec:dataconstruction}
\begin{figure}[!t]
\centering%
    \includegraphics[width=\linewidth]{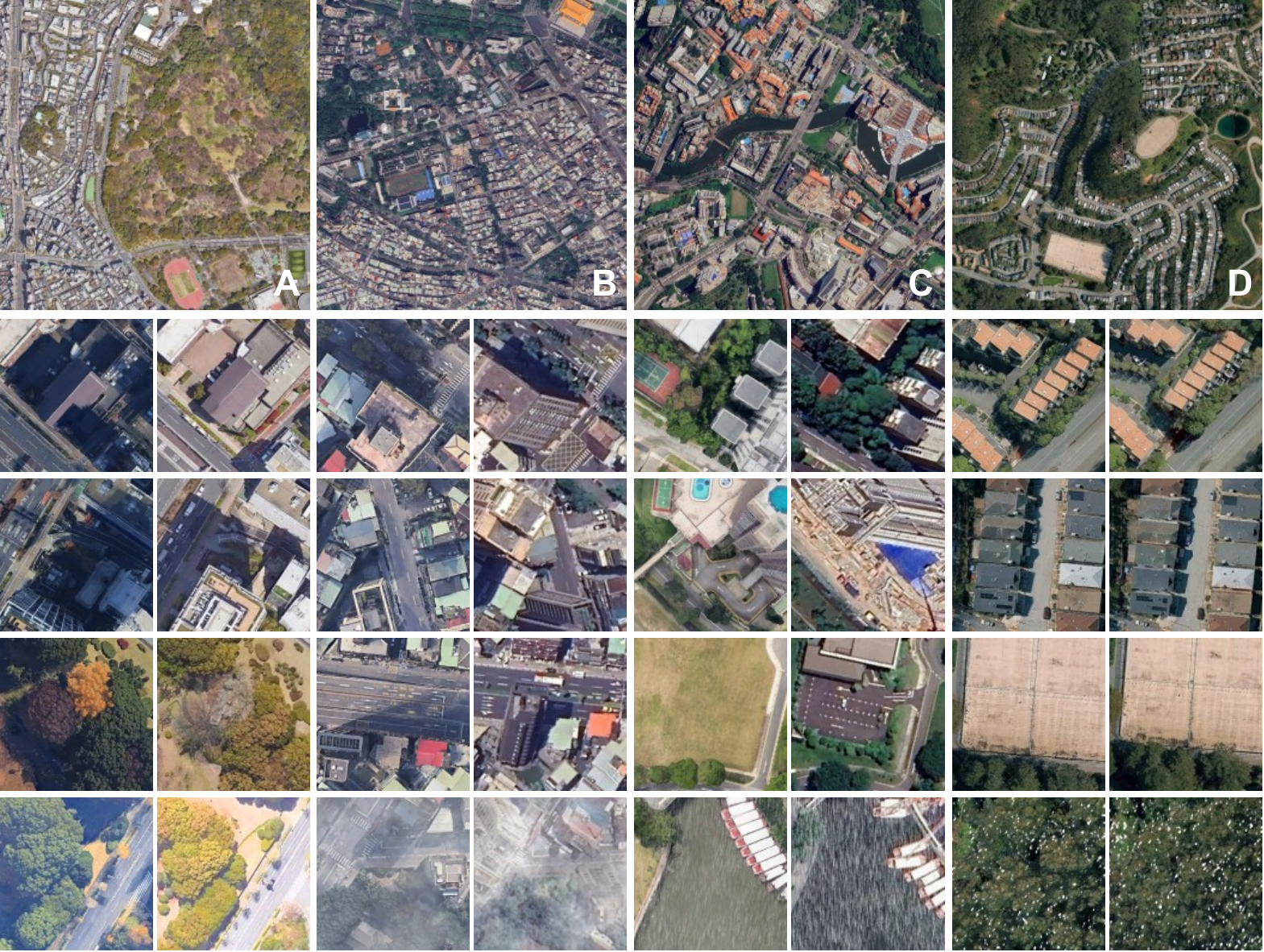}
\caption{Satellite images of four cities with distinct landscapes. Below the satellite images are their selected U-S cross-view image pairs (UAV views left). City~A is dominated on the right side by vegetation; City~B consists mainly of densely packed low-rise buildings; City~C contains many tall buildings and a wide river; City~D lies in mountainous terrain with sparse, highly similar buildings. They exhibit prominent cross-view appearance gaps, illumination variations, and scene changes, covering diverse landforms such as buildings, roads, forests, mountains, and rivers.}%
\label{fig:city1234}%
\end{figure}

We construct our dataset from four cities, each represented by a square, contiguous satellite image, with details illustrated in~\cref{fig:city1234}. We first describe the data collection process, followed by the dataset structure.

\paragraph{Sampling Process}
For each city, we first download the entire satellite image in satellite-view and then extract cross-view samples block by block from the corresponding remote-sensing blocks (RSBs). 
\begin{figure}[!t]
\centering%
    \includegraphics[width=\linewidth]{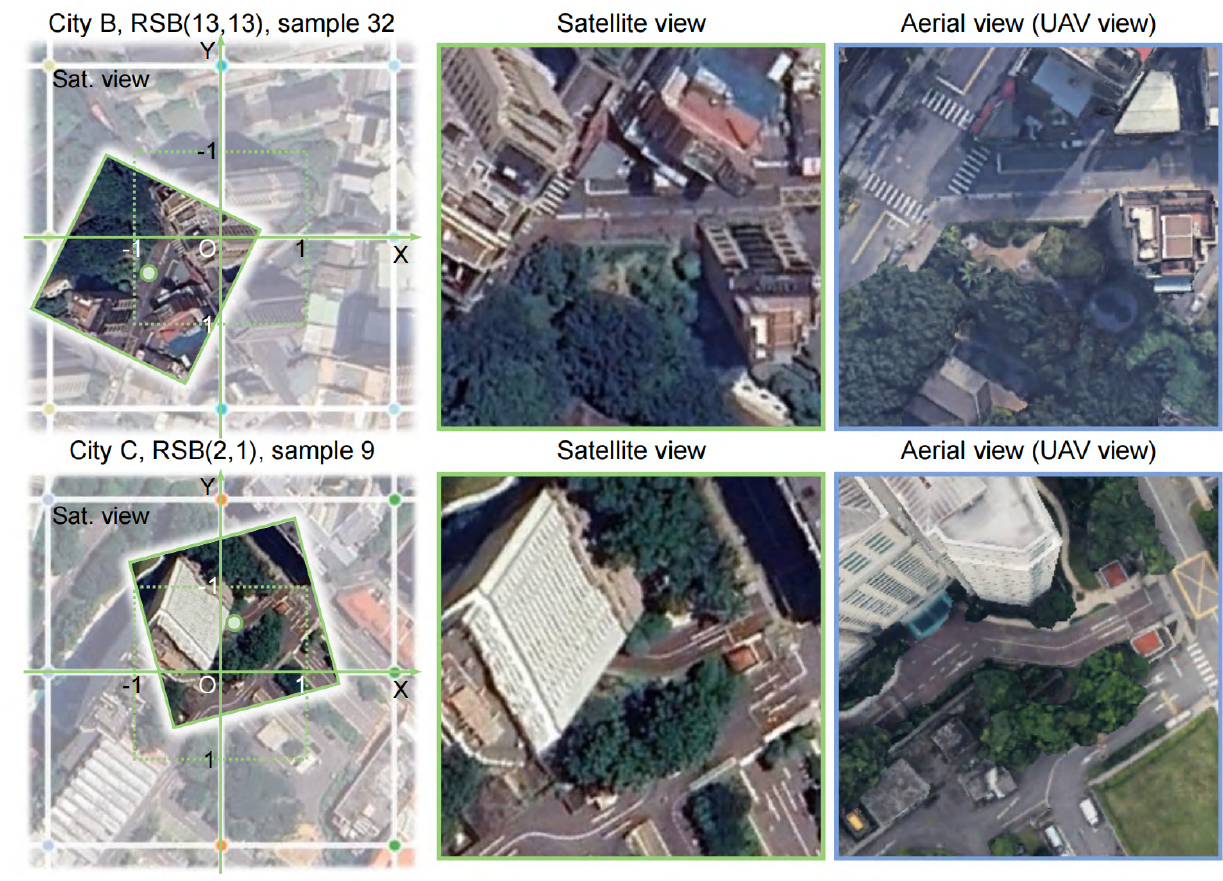}
\caption{Two examples of UAV-satellite cross-view UVP sampling during dataset construction. 
In each row, the left panel shows the sampled patch in the RSB, and the middle and right panels show the corresponding satellite-view patch and UAV-view patch.}%
\label{fig:supsample}%
\end{figure}
As illustrated in~\cref{fig:supsample}, the top row shows an example RSB in City B indexed by $(13,13)$. The X-axes and Y-axes passing through the block center partition this RSB into four adjacent RSTs. 

The process of constructing cross-view samples can be summarized in the following two steps:
\textbf{(1)} In Google Earth 2D mode (satellite-view mode), within the effective sampling area (green dashed box) defined in the local $XOY$ coordinate system of the RSB, we randomly sample relative coordinates and headings to obtain a satellite-view patch (the 32nd sample). The selected patch, shown as the dark patch with a green border centered at the green dot (the top-middle panel), serves as an ideal reference.
\textbf{(2)} Using the same geodetic coordinates and heading angle, we then query Google Earth 3D mode (UAV-view mode) and extract the corresponding UAV-view patch (UVP), as illustrated by the blue-bordered image in the top-right panel. 

The second row shows an analogous process for the RSB indexed by $(2,1)$ in City C, where we obtain the four RSTs together with the 9th satellite-view patch and its paired 9th UVP. 
Unlike satellite-view, the UAV-view mode in Google Earth renders a photogrammetry-based 3D reconstruction of the entire city. 
All buildings, vegetation, and terrain are represented as textured 3D meshes, enabling oblique viewpoints that closely approximate what a real UAV camera would observe at the same position. 
Consequently, UVPs provide a more realistic approximation of true UAV perspectives compared with satellite-view patches.
 
\paragraph{Basic Composition of the Dataset}
Following this procedure, we sample $100$ cross-view pairs from every RSB. Each city's image contains 15$\times$15 overlapping RSBs whose effective sampling areas are designed to seamlessly tile the satellite image (except for an outer margin of 128\,pixels that cannot be fully covered). As a result, the four city images yield a total of 90k samples per view (i.e., 90k satellite-view patches and 90k UVPs).

For each sample, the basic metadata include one UVP (paired with the corresponding satellite-view patch), its relative coordinates and heading angle (encoded as a cosine-sine vector), the four adjacent RSTs which are shared by all samples in the same RSB block, and the file paths to all associated image tiles. These fields are stored in a single \texttt{metadata.csv} file, which can be directly used for training and evaluation. 
As part of the dataset, \texttt{metadata.csv} provides the paths to the corresponding entity images (all UVPs and RSTs) during training and testing.

In addition, every satellite image, and UVP is associated with a JSON file that records auxiliary information such as the image name, center latitude and longitude, pixel resolution, spatial extent, camera height, tilt angle, and basic conversion factors, ensuring that the dataset is well documented and easily reusable.

\begin{figure}[!t]
\centering%
    \includegraphics[width=\linewidth]{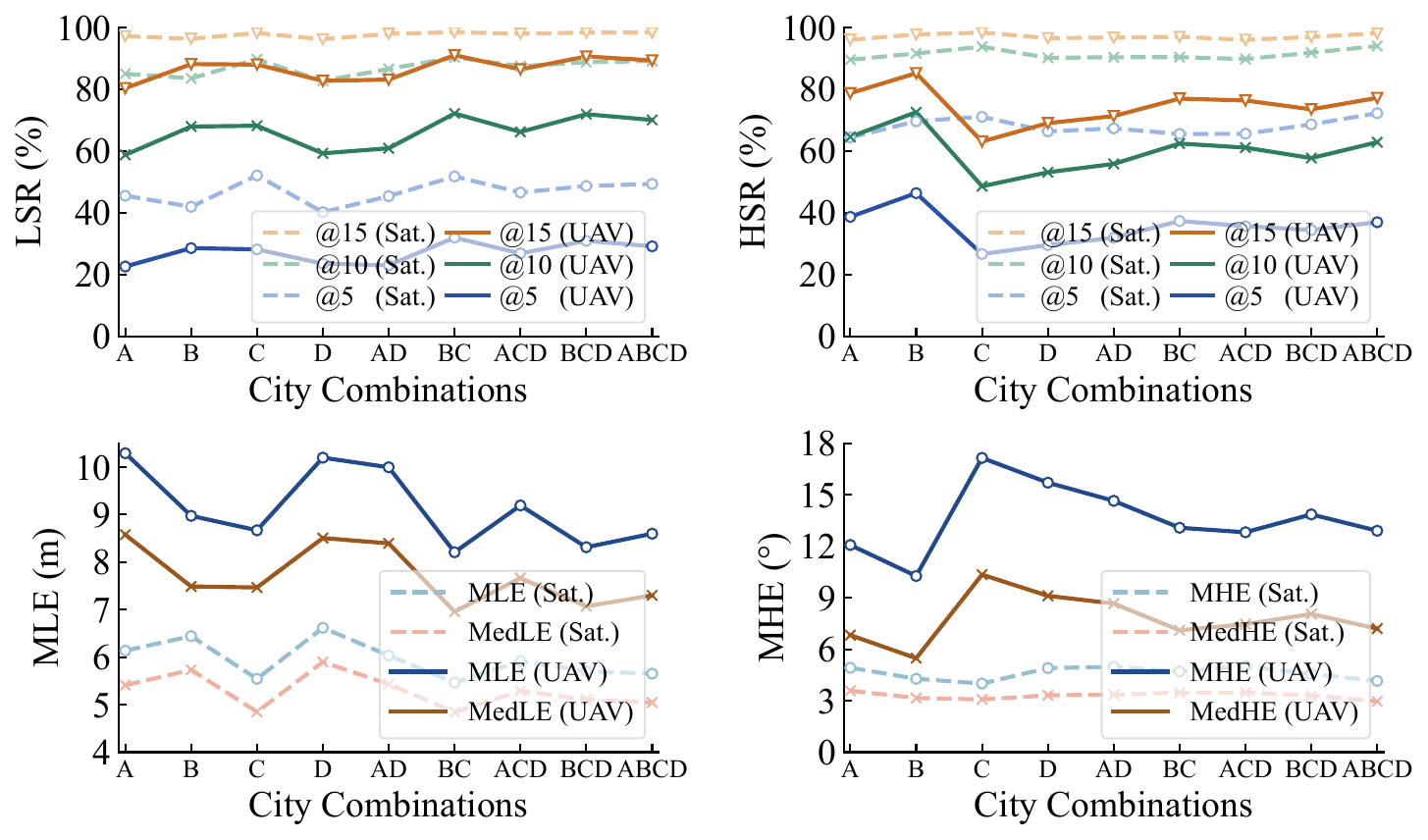}
\caption{Localization and orientation performance under different city combinations.}%
\label{fig:multicitycurve}%
\end{figure}

\subsection{Impact of Different City Combinations}
\label{sec:dsetscale}

We employ more fine-grained metrics to evaluate the effect of city combinations on UAV bearing estimation. 
As shown in ~\cref{fig:multicitycurve}, the combinations include four single-city settings, two two-city pairs, two three-city mixtures, and one four-city configuration that exposes the model to the entire set. 

The models trained on single-city subsets exhibit noticeable differences in localization and heading performance on their corresponding satellite images, especially in the UAV-view setting.
This suggests that variations in urban morphology have a non-negligible impact on purely vision-based localization and navigation.

Specifically, City B achieves relatively strong overall performance: it features a dense, nearly grid-like street network and a large number of small buildings, providing rich fine-grained structural information and strong, consistent cues for both localization and orientation.
Although City C also contains many buildings that offer abundant structural texture, the tall high-rise buildings induce larger cross-view appearance discrepancies, and the presence of a long river corridor with relatively sparse visual features is likely a major factor behind the degradation in heading accuracy.
City A and City D show comparatively worse localization and heading performance, which correlates with the presence of large green areas, mountainous terrain, and many visually similar buildings.

From a global perspective, a particularly noteworthy trend is that the LSR and MLE do not degrade when the number of cities increases and the sample diversity grows; instead, they exhibit a clear upward trend.
Meanwhile, the heading success rate and heading accuracy remain roughly at an average, stable level.
This indicates that, as more cities are included and the training data become more diverse, \method is able to learn more generic orientation cues that generalize across heterogeneous city layouts, rather than overfitting to any single satellite image.

\section{Effect of Weather Augmentation}
\label{sec:weatheraug}

\begin{figure}[!t]
\centering%
    \includegraphics[width=\linewidth]{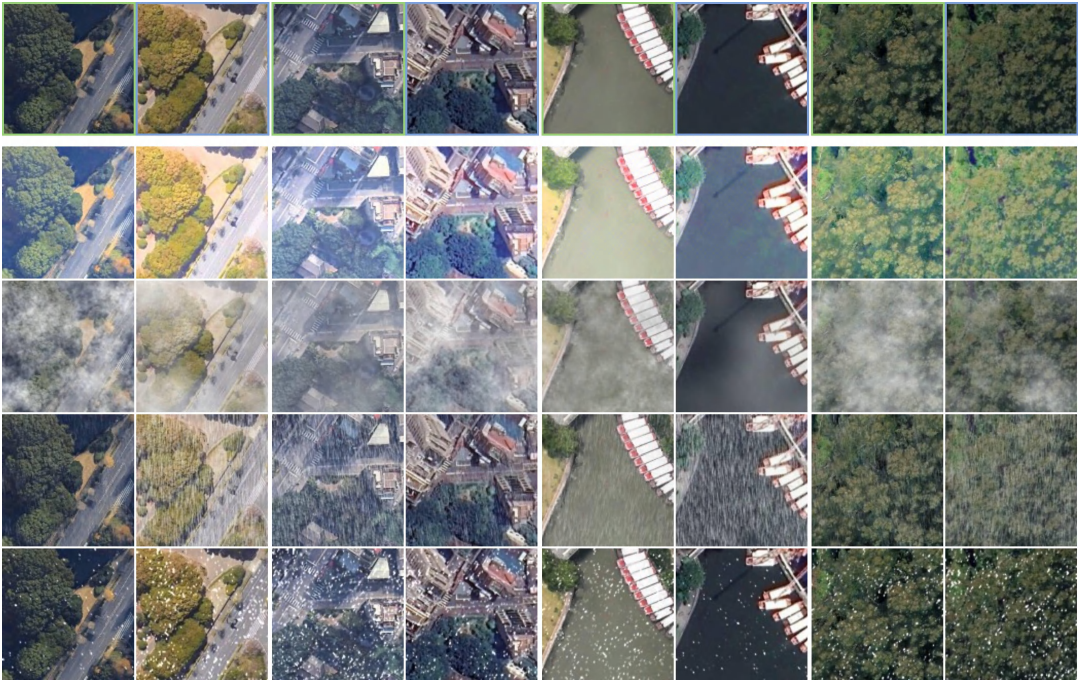}
\caption{Four types of weather augmentations. The first row shows four original cross-view sample pairs from the four cities, where the green and blue boxes denote the UVP and its corresponding satellite-view patch. The four rows below visualize the corresponding illumination, fog, rain, and snow augmentations.}%
\label{fig:supweatheraug}%
\end{figure}
\begin{figure}[!t]
\centering%
    \includegraphics[width=\linewidth]{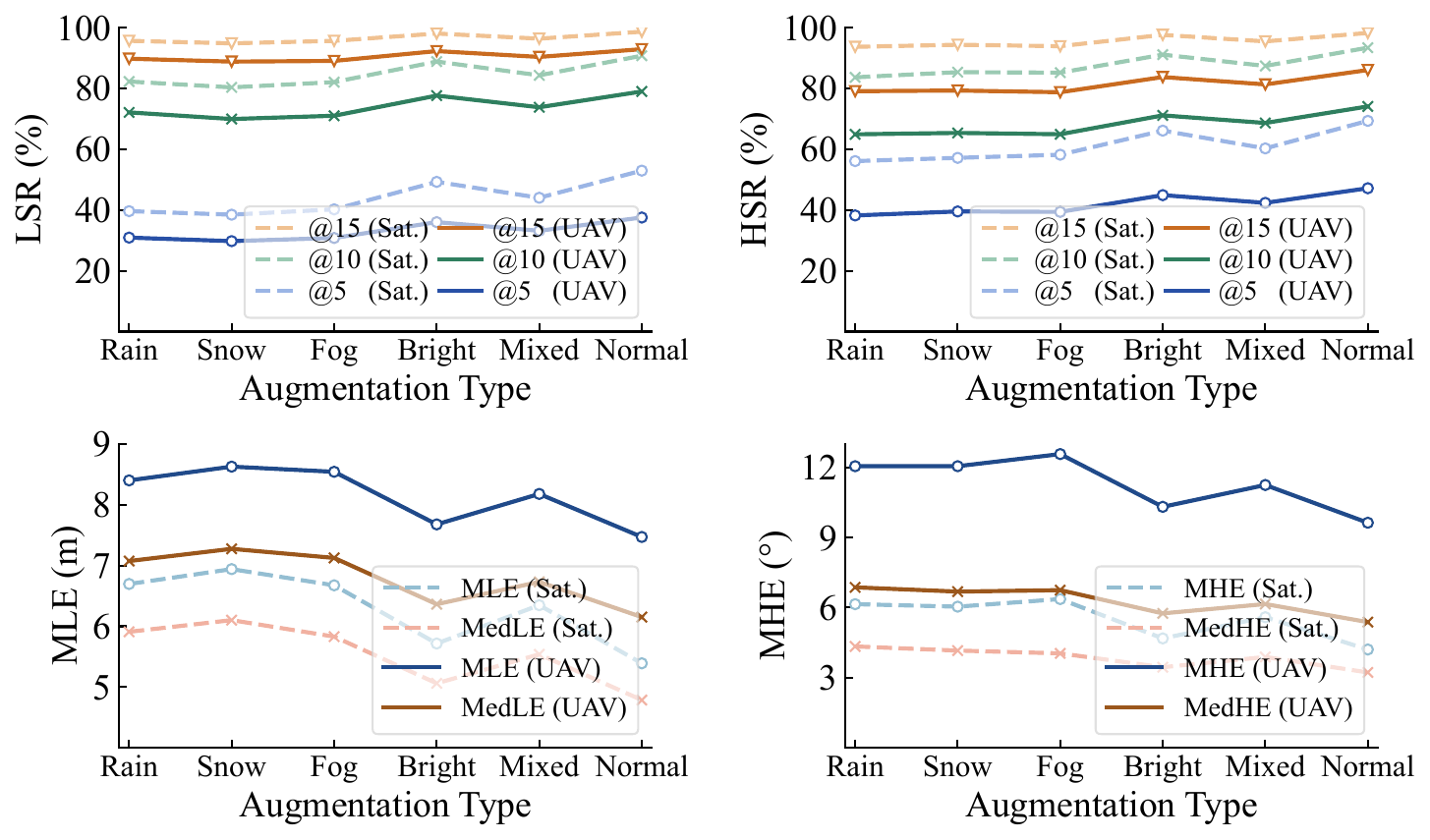}
\caption{Effect of Weather Augmentation on \method. Mixed-weather training further boosts and stabilizes performance compared with the non-augmented setting.}%
\label{fig:supweathercurve}%
\end{figure}

We apply weather augmentation to 20\% of the training samples for each of the four weather types and evaluate the augmented model under six test settings (four single-weather conditions, a mixed-weather condition, and a no-augmentation normal condition). Four types of weather augmentation are shown in~\cref{fig:supweatheraug}, and the fine-grained metric curves are shown in~\cref{fig:supweathercurve}.

Overall, the augmented model consistently improves both success rate and localization/heading accuracy on unseen samples. More importantly, the performance curves exhibit similar trends across different weather settings, indicating that the model learns a shared, weather-robust representation rather than overfitting to a specific appearance pattern.

Among the four types, illumination augmentation provides the most significant gain. Considering the large brightness discrepancies between UAV and satellite views, this suggests that illumination augmentation effectively mitigates cross-view appearance gaps and is particularly beneficial for cross-view localization and navigation.

\section{Visual Analysis of Localization and Orientation Performance}
\label{sec:locerrstat}

To more comprehensively analyze the factors that affect the localization and orientation performance of \method at different levels, and to provide more intuitive support for this analysis, we perform multi-granularity visualization and statistical analysis of the evaluation metrics at three levels: city-level, RSB-level, and sample-level.

\subsection{City-level Statistical Analysis}
\begin{figure*}[!t]
\centering%
    \includegraphics[width=\linewidth]{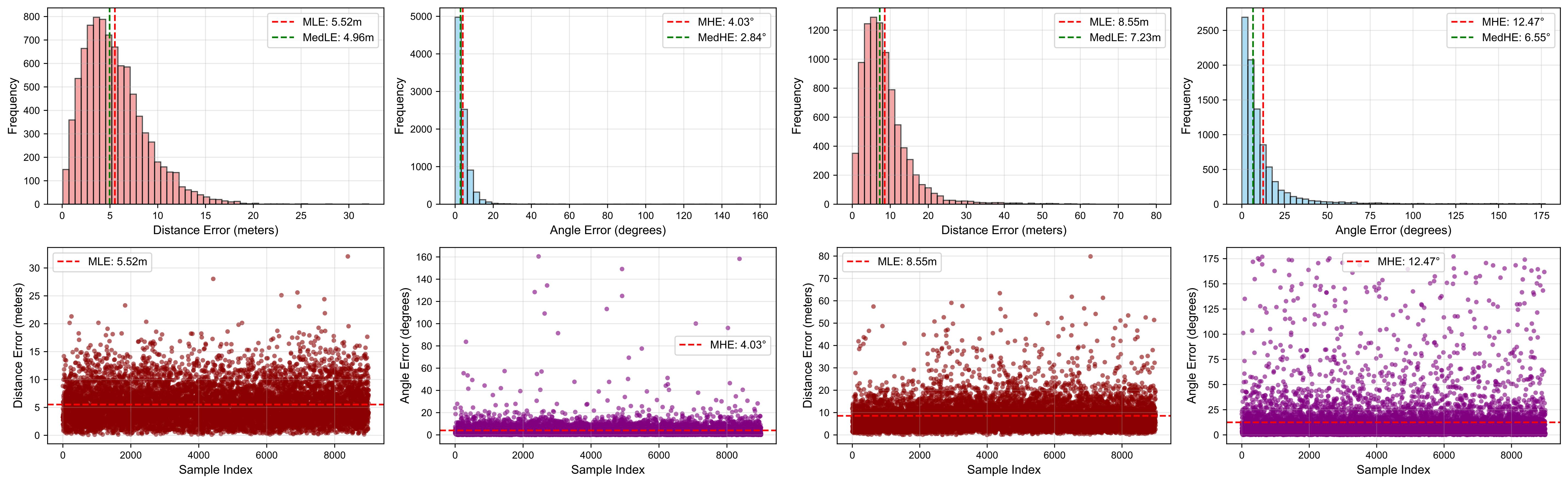}
\caption{Localization and heading error distributions and scatter plots of 9k unseen test samples in satellite and UAV views.
}%
\label{fig:locerrstat}%
\end{figure*}

For the VGG-16 backbone version of \method, the localization error (LE) and heading error (HE) distributions and scatter plots for the 9k unseen test samples (accounting for 10\% of all samples) in the satellite and UAV views are shown in~\cref{fig:locerrstat}.

The two left columns show the localization and heading errors in the satellite view, while the two right columns correspond to the UAV view.
The first row presents the error distributions, and the second row shows the error scatter plots.
We can first observe that, in both views, the localization and heading errors follow a long-tailed distribution that is highly concentrated around small values, with the median errors consistently lower than the corresponding means.
This effect is particularly pronounced for the heading errors in the UAV-view setting.
Moreover, compared with the satellite view, the UAV view (U-S cross-view) setting exhibits noticeably more outliers, which is closely related to the increased complexity of cross-view localization.

In the bottom row, the scatter plots show that the vast majority of samples stay close to the horizontal median-error lines, and large-error outliers are relatively sparse and dispersed across the sample index, without obvious structural bias or drift.
Distance errors rarely exceed 20--25\,m, whereas heading errors exhibit a few extreme outliers above $45^\circ$, suggesting that heading estimation is more susceptible to rare failure cases than localization.
Overall, ~\cref{fig:locerrstat} indicates that \method achieves stable and accurate localization and orientation on most unseen samples in both views, with only a small fraction of challenging cases contributing to the long tails of the error distributions.

\subsection{RSB-level Statistical Analysis}

During our evaluation, we observed that the accuracy of UAV localization and orientation is significantly influenced by terrain, and that this effect differs between the satellite view and the UAV view.

\begin{figure}[!t]
\centering%
    \includegraphics[width=\linewidth]{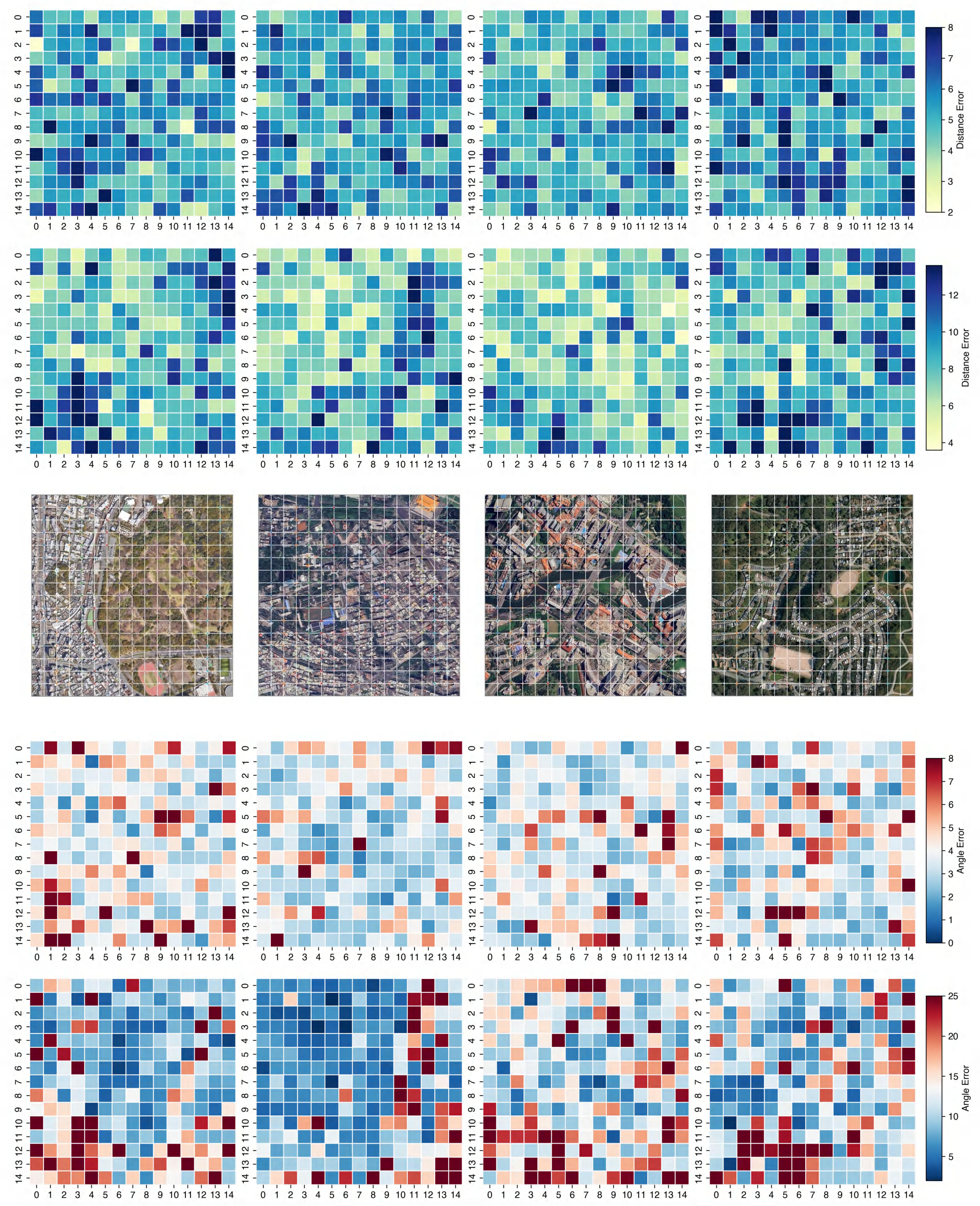}
\caption{Heatmaps of mean localization and heading errors aggregated per RSB in two views across four cities.}%
\label{fig:hot}%
\end{figure}

To analyze this phenomenon, we compute the mean localization and heading errors for each RSB and visualize them as RSB-level heatmaps, as shown in~\cref{fig:hot}, which depict the spatial distribution of the model's accuracy across different terrain types.
From top to bottom, the rows show the RSB-level MLE heatmaps in the satellite view and UAV view, satellite imagery of the four cities, and the RSB-level MHE heatmaps in the satellite view and UAV view.
From left to right, the four columns correspond to City A--D.
To enhance the visualization, we center the color scale of each heatmap at the mean error and set a symmetric range around this mean value.

We mainly analyze the relationship between terrain and errors under the UAV-view setting.
From the RSB-level localization error heatmaps, we observe that dark-blue regions in the lower-left of City A, the right side of City B, and light-blue region in City C mostly correspond to tall building areas; these regions also exhibit relatively large heading errors, indicating that strong cross-view perspective changes increase both localization and orientation errors.
In contrast, the central-upper area of City B and the central area of City A show lower localization and heading errors; these regions are dominated by low-rise buildings and small green spaces with rich fine-grained structural textures and small appearance differences, which are favorable for accurate localization and orientation.
By comparison, the upper-right region of City A and the lower-left region of City D are mainly covered by forests or open fields, where sparse features make localization and orientation more difficult.
Regions with highly similar building appearances can also cause degrade performance, such as the plaza on the central-right of City C and the building cluster in the upper-right of City D.

In addition, the satellite-view heading-error heatmap of City C shows that the central river is associated with larger heading errors, consistent with the fact that rivers provide very limited texture cues.
Moreover, since UVP samples are collected in different seasons and at different times than the satellite imagery, illumination changes, color shifts, and scene changes also affect localization and orientation.
Comparing the two views, the spatial distributions of large errors vary, reflecting cross-view appearance gaps, especially perspective distortion around tall buildings, illumination discrepancies, and changes in surface objects.

Overall, these observations confirm that large cross-view visual discrepancies, regions with sparse structural textures, and areas with highly similar appearances all have a significant influence on the performance of the proposed algorithm.

\subsection{Sample-level Visual Analysis}
\begin{figure*}[!t]
\centering%
    \includegraphics[width=\linewidth]{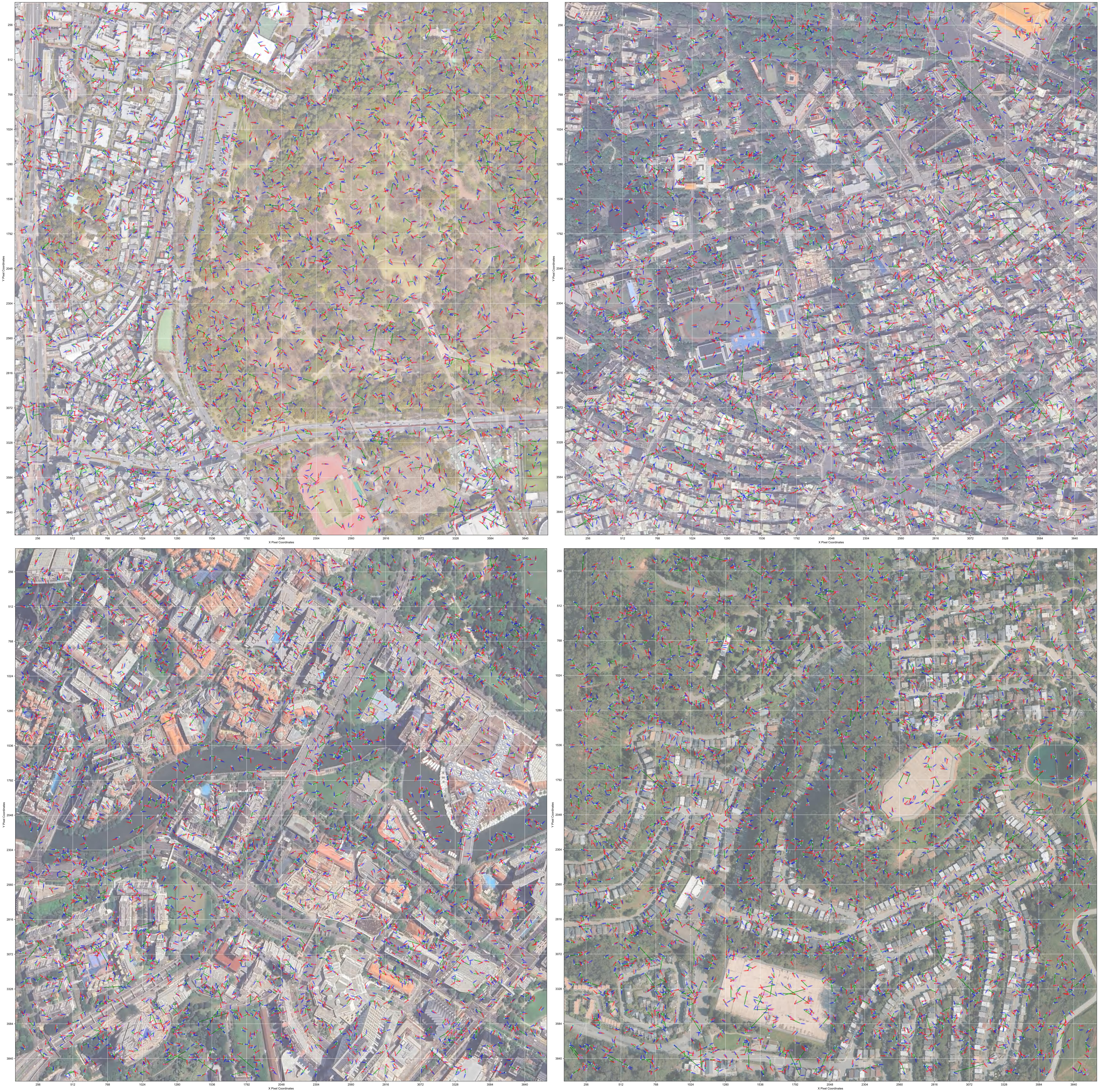}
\caption{Visualization of U-S cross-view localization and heading errors on real satellite images from four cities.
The four satellite images in the \textbf{top-left, top-right, bottom-left, and bottom-right panels} correspond to \textbf{City A--D}, respectively, and overlay the localization and heading results of all unseen test samples under cross-view setting. 
In each image, \textit{blue/red dots} denote the ground-truth/predicted positions, while \textit{blue/red arrows} denote the ground-truth/predicted heading vectors. 
The \textit{green line segment} connecting the endpoints of the two arrows represents the position error, whose magnitude is annotated by \textit{green numerical values}, and the \textit{purple numeric values} denote the heading error; smaller values indicate better accuracy, and perfect overlap between the red and blue arrows corresponds to the ideal case. 
This visualization enables precise, fine-grained inspection of localization/heading errors under different scene types across all cities. 
The satellite images are divided into fine-grained RSTs, such that the surrounding terrain of any sample can be examined and analyzed in detail against the real-world surroundings.
For example, in the high-rise building areas (\textit{e.g.}, the lower-right of City~A and the lower region of City~C), in feature-sparse regions (\textit{e.g.}, forests in the upper-right of City~A, green fields and the central river in City~C, and open grounds in City~D), as well as in dense clusters of similar buildings (\textit{e.g.}, City~B and City~D), many samples exhibit noticeably large localization and/or heading errors.
These errors are likely due to the adverse effects of cross-view appearance discrepancies, feature sparsity, and feature similarity.}%
\label{fig:supvecmap43d}%
\end{figure*}

To support the analysis from a finer-grained perspective, we perform map-based sample-level visualization of localization and heading errors under U-S cross-view, revealing detailed error patterns and enabling intuitive visualization of localization errors (LE) and heading vectors (HE). For each of the four cities, we visualize LE and HE across 2250 unseen samples overlaid on corresponding satellite images, as shown in~\cref{fig:supvecmap43d}. This geographic visualization helps clearly illustrate how errors vary across different urban regions. To facilitate reading and cross-referencing with the figure, we include the relevant details in the caption for clarity.


\section{More Navigation Experiments}
\label{sec:navtest}
\begin{figure*}[!t]
\centering%
    \includegraphics[width=\linewidth]{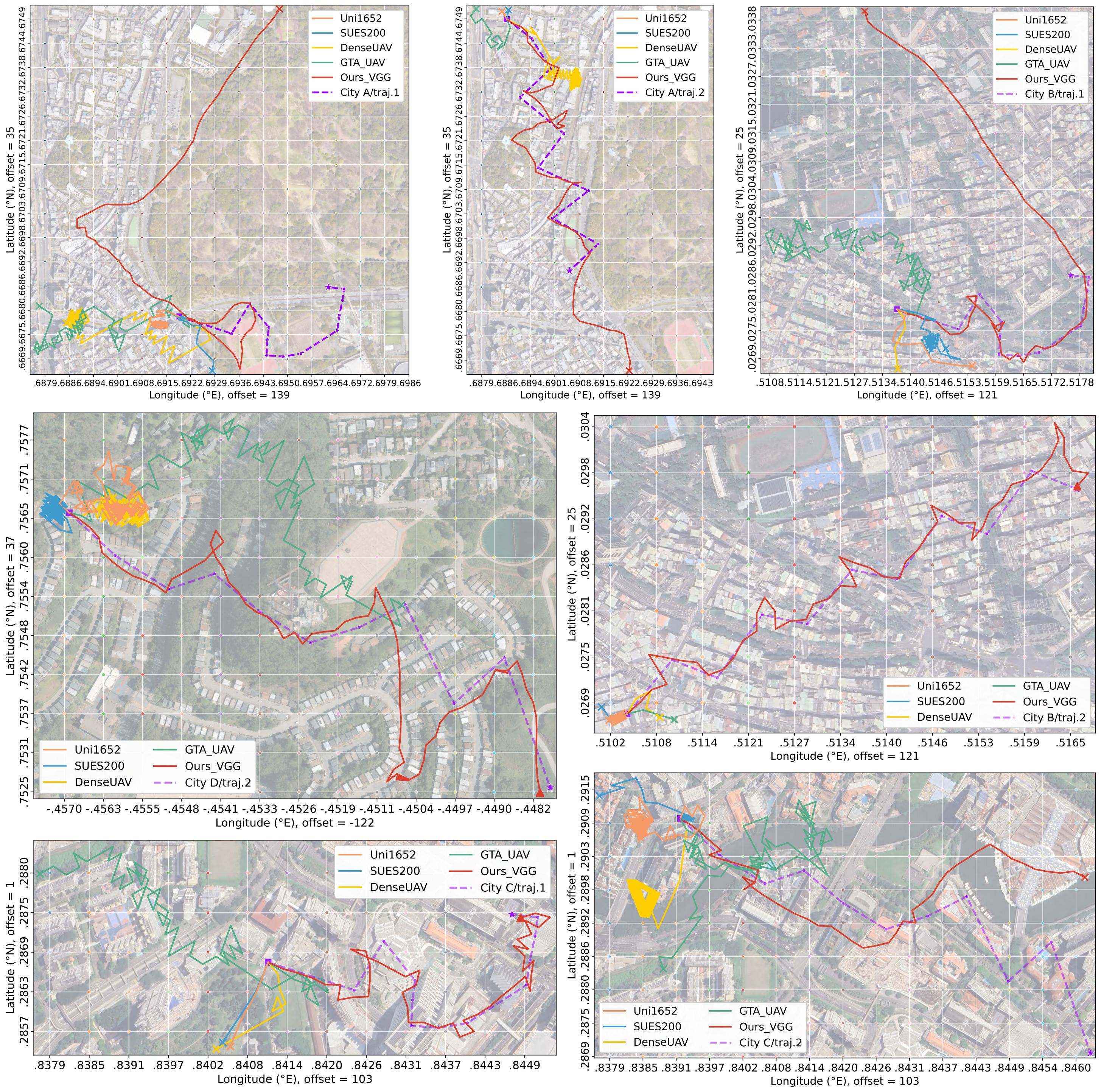}
\caption{Navigation performance comparison on seven additional routes.
We further provide a comprehensive comparison between \method (red trajectories) and four baseline methods on seven additional navigation routes across the four cities. 
The UAV step size is 25\,m, and the waypoint arrival threshold is 20\,m. All visual elements follow those in Fig.~6 of the main paper. 
The eight routes across Cities A--D range in length from 524 m to 1119 m.  
Each route contains 10--13 waypoints and covers diverse scene types, including buildings of varying scales, green areas, rivers, and playgrounds.
Overall, \method exhibits the strongest adaptability to different urban layouts and highly winding paths. 
It successfully completes four of the eight routes and, for the remaining ones, still follows most of the designated path before failure.  
In contrast, baseline methods struggle in these complex scenes and along curved routes. 
The green GTA-UAV trajectories are often longer but show significant deviation from the reference path. 
The orange Uni-1652, yellow DenseUAV, and blue SUES-200 trajectories tend to drift away near the starting point, with some falling into local loops and circling within a small region.
Notably, large positional deviations and flight drift mainly occur in vegetation areas (traj. \#1 City A, traj. \#2 City A, traj. \#2 City D), similar building areas (traj. \#1 City B, traj. \#2 City D), and tall building areas (traj. \#2 City C).
These regions often suffer from feature sparsity, high feature similarity, and significant cross-view appearance discrepancies, all of which degrade localization and heading estimation performance.}%
\label{fig:navtraj7}%
\end{figure*}

We further comprehensively compare \method (red trajectories) with four baselines on seven additional navigation routes across four cities, as shown in~\cref{fig:navtraj7}.
The eight routes (City A--D, Route \#1 and \#2) have lengths of 771, 1119, 644, 757, 524, 821, 721, and 1115\,m, respectively, with 10--13 waypoints each and diverse scene types.
While not covering the full satellite image, these routes span a wide range of scene types and scales (\textit{e.g.}, large/small buildings, similar building clusters, roads, rivers, open fields, forests, and playgrounds), thus reflecting the model’s navigation performance under different scene layouts.

\subsection{Analysis of Multiple Flight Trajectories}

From these visualizations, \method shows the strongest adaptability to multi-city scenarios and complex navigation routes, consistently achieving stable performance across diverse scenes, such as dense urban blocks, irregular river boundaries, and mixed residential--commercial areas.
It successfully completes four routes and, on the remaining ones, still follows most segments before failure. 
These failures typically occur at later stages and are mainly caused by severe appearance ambiguities or rapid structural changes, rather than early-stage drifts. 
This suggests that \method maintains stronger long-horizon consistency under cross-view misalignment.

In contrast, baseline methods struggle in structurally complex scenes or along routes with tight curves and multi-directional turns. 
GTA-UAV (green) trajectories are often longer but frequently diverge from the intended path and continue advancing in an incorrect direction, indicating unstable behavior. 
Uni-1652 (orange), DenseUAV (yellow), and SUES-200 (blue) exhibit more severe issues, often drifting shortly after the starting point. 
In several cases, they fall into local loops, circling within a small region instead of progressing along the intended route.

\subsection{Analysis of Flight Frame Sequences}
\begin{figure*}[!t]
\centering%
    \includegraphics[width=\linewidth]{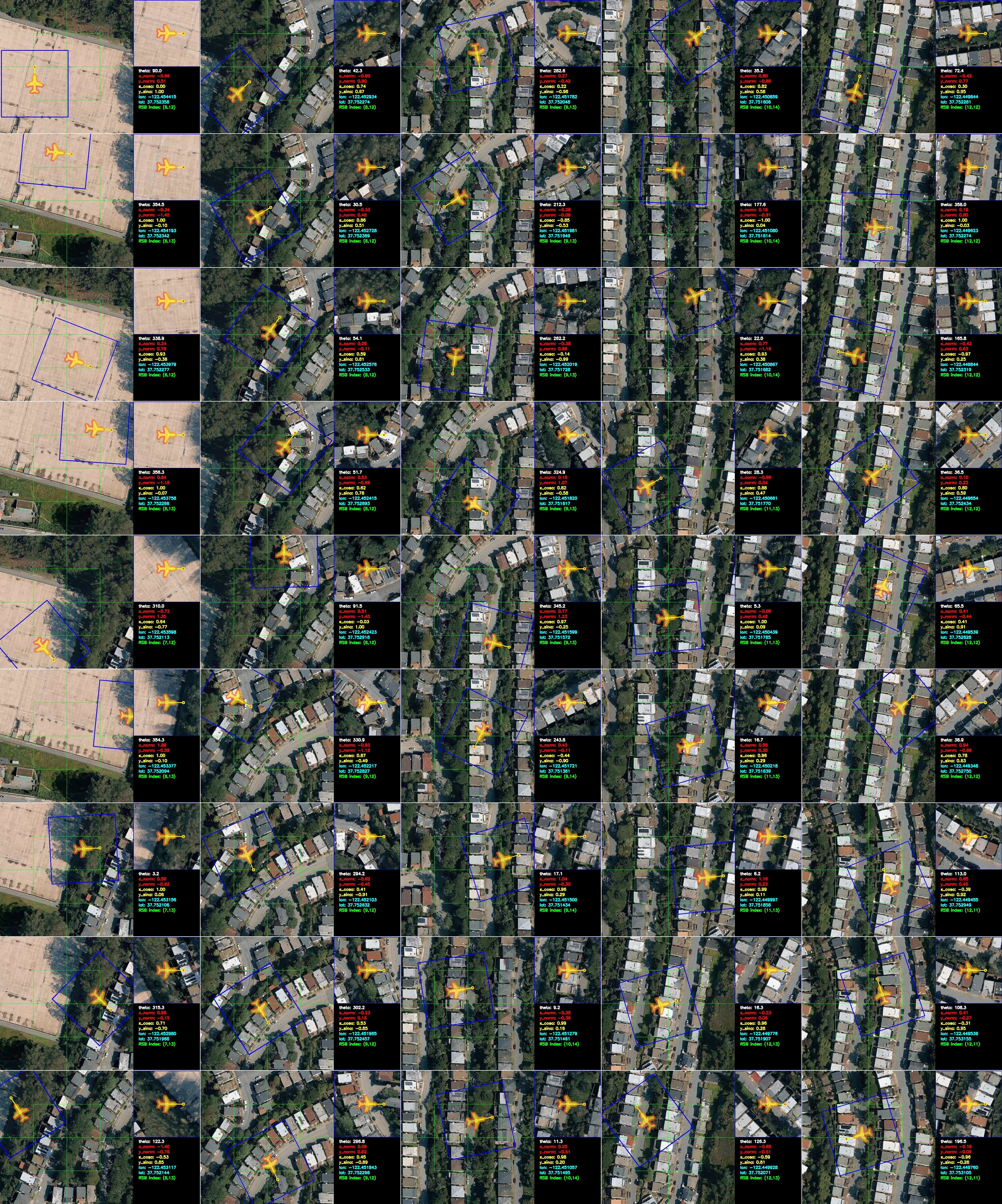}
\caption{Visualization of cross-view navigation frames of \method.
We present a successful case for trajectory~\#1 in City~D from Fig.~6 of the main paper.
Frames are arranged top-to-bottom and left-to-right. 
In each step, the left panel shows the current RSB, where the blue rectangle indicates the UAV field of view and the airplane icon indicates the heading. 
The top-right inset shows the UVP, and the bottom-right inset reports outputs including heading angle, relative coordinates, direction vector, geographic coordinates, and the RSB index.
}
\label{fig:ourvggd50frames3d}%
\end{figure*}

\begin{figure*}[!t]
\centering%
    \includegraphics[width=\linewidth]{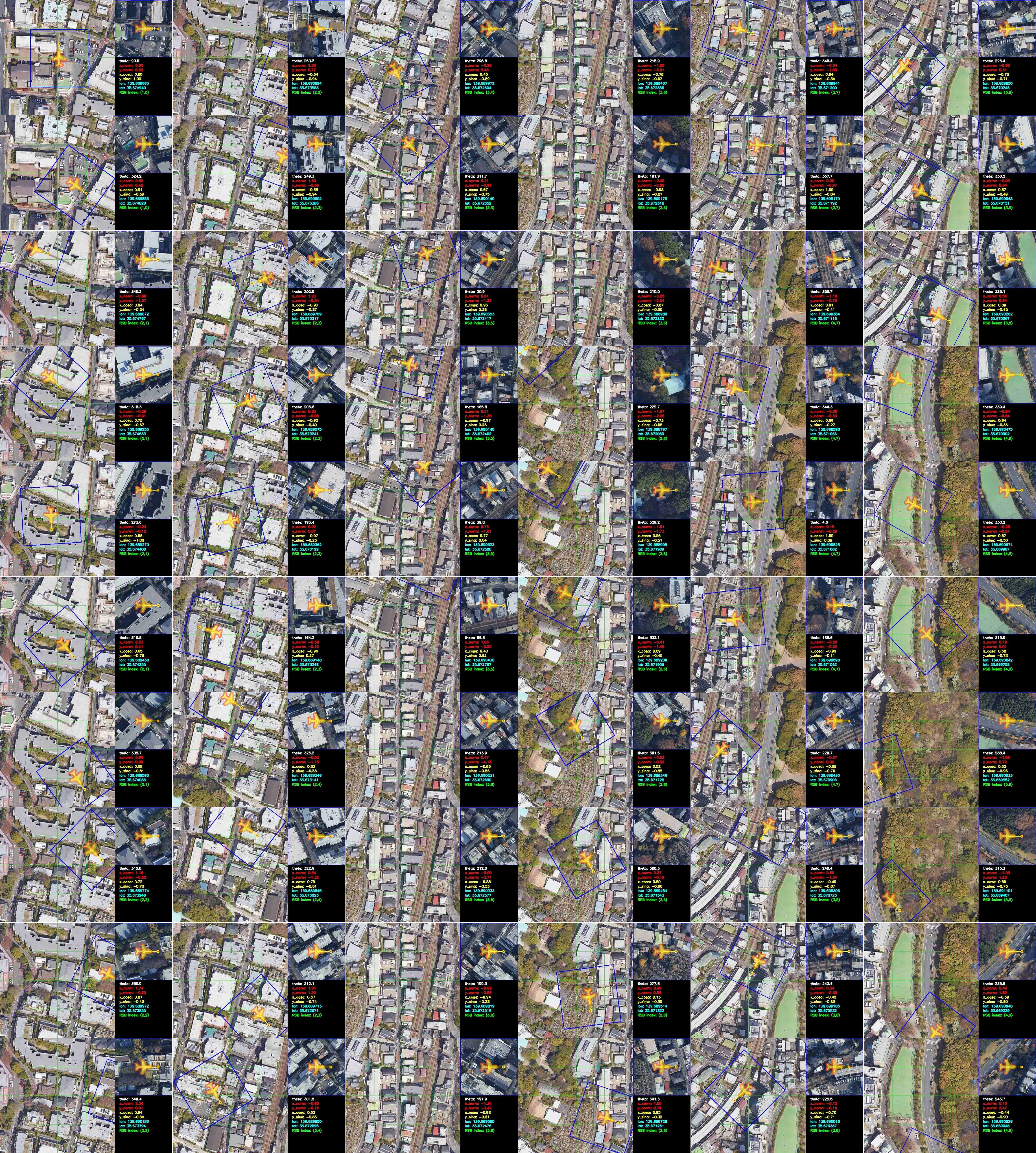}
\caption{The UAV cross-view navigation process corresponding to trajectory~\#2 in City~A, as shown in~\cref{fig:navtraj7} (first row, middle).
This is a more challenging case in which navigation eventually fails. However, the UAV still follows a substantial portion of the planned path before drifting away. 
We visualize overlapping portion (first 60 steps) to highlight cross-view discrepancies. 
Significant spatial differences between UVP and satellite patches arise from 3D parallax and illumination variations at low altitudes. 
Deviations occur at frames 26--34, likely due to waypoint turns and repetitive building patterns. 
Despite these challenges, \method remains robust along winding trajectories.}%
\label{fig:ourvgga51frames3d}%
\end{figure*}

To provide an intuitive illustration of the cross-view UAV navigation process, we visualize step-by-step trajectories for two representative routes, including the sequence of visited satellite-view RSBs and corresponding UVP frames, enabling analysis of how the system progressively adjusts heading, updates relative position, and aligns with the target route under varying scene structures and cross-view discrepancies (see~\cref{fig:ourvggd50frames3d,fig:ourvgga51frames3d}).

Corresponding to trajectory~\# 1 of City D in Fig.~6 of the main paper, \cref{fig:ourvggd50frames3d} shows a successful navigation case, in which \method completes the route in 45 steps while maintaining stable alignment with the designated trajectory.
Frames 10--14 and 30--35 show smooth transitions. 
Noticeable heading changes occur at frames 8--9, 14--15, 24--25, 35--36, and 44--45, which are near waypoints and correspond to normal turns, demonstrating that the model can effectively handle turning maneuvers.
In contrast, consecutive large heading changes appear at frames 28--30 and 38--40. 
At these positions, the field of view contains repeated patterns (\textit{e.g.}, rows of similar buildings and vegetation), indicating that feature similarity or repetition can introduce stochastic disturbances to heading estimation.

\cref{fig:ourvgga51frames3d} shows a more challenging case from trajectory~\#2 of City A, where navigation eventually fails, but the UAV still manages to follow a substantial portion of the planned path before drifting away. 
Together, these visualizations offer insights into the model’s decision-making behavior, the challenges caused by viewpoint inconsistencies, and the contrast between successful and near-successful trajectories.

\section{More Ablation Experiments}
\label{sec:ablation}

We also take ablation study under the satellite-view settings.
As shown in~\cref{tab:table_supablation}, the satellite-view results follow the same trend as in the UAV view: adding the GLUF module yields a clear performance gain in both views.
Moreover, the RCE and PSG modules further enhance localization and heading accuracy, with particularly strong improvements in heading estimation. 
The RCE, PSG, and CA exhibit slightly better orientation than localization performance.

\begin{table}[!t]
\centering
\setlength{\tabcolsep}{1.3pt}      
\renewcommand{\arraystretch}{1.0}   
\caption{Ablation study under satellite-view settings.}
\label{tab:table_supablation}
\begin{tabular}{@{}llllrrrrr@{}}
    \toprule
    \textbf{GLUF} & \textbf{RCE} & \textbf{PSG} & \textbf{CA}
    & \textbf{R@1}\textsuperscript{\scriptsize$\uparrow$} 
    & \textbf{LSR}\textsuperscript{\scriptsize$\uparrow$}
    & \textbf{HSR}\textsuperscript{\scriptsize$\uparrow$}
    & \textbf{MLE}\textsuperscript{\scriptsize$\downarrow$}
    & \textbf{MHE}\textsuperscript{\scriptsize$\downarrow$} \\
    \midrule
    
    \ding{55} & \ding{51} & \ding{51} & \ding{51} & 67.33 & 58.56 & 83.67 & 14.55 & 10.41 \\
    \ding{51} & \ding{55} & \ding{51} & \ding{51} & 90.55 & 98.10 & 95.11 &  5.74 &  5.51 \\
    \ding{51} & \ding{51} & \ding{55} & \ding{51} & 90.65 & \textbf{98.40} & 96.96 &  5.70 &  4.72 \\
    \ding{51} & \ding{51} & \ding{51} & \ding{55} & 89.88 & 98.28 & 96.30 &  5.68 &  5.12 \\
    \ding{51} & \ding{51} & \ding{51} & \ding{51} & \textbf{90.76} & 98.33 & \textbf{98.12} &  \textbf{5.65} &  \textbf{4.15} \\
    
    \bottomrule
\end{tabular}
\end{table}

\section{Limitations}
\label{sec:suppl_lim}
Vision-only UAV localization remains an open problem and is still far from mature.
Vision-only performance is a critical complement to sensor-vision fusion under GNSS denial or long-term inertial drift, and also underpins geometric reasoning methods (\textit{e.g.}, PnP) that estimate multi-DoF UAV poses. 
However, M2T/retrieval paradigms are limited by grid density and typically ignore heading estimation, restricting long-range localization and autonomous navigation. 
Moreover, repeated encoding of satellite tiles makes M2T-based methods time-consuming even when with onboard satellite imagery.
In contrast, our paradigm avoids this cost and achieves state-of-the-art vision-only localization while jointly estimating heading, enabling more complete conditions for long-distance autonomous navigation.

Nevertheless, although our method focuses on challenging scenarios such as misalignment, sparse features, and varying IoUs, real-world flight environments involve many additional temporal and spatial variations. Thus, our approach still has several limitations. 
First, its generalization to unseen cities remains underexplored. The model is trained and evaluated on satellite images from four cities, where supervised learning allows effective regression of position and heading. However, its cross-city transfer ability requires further validation and improvement. Second, the current framework does not explicitly address dynamic scene changes, such as moving vehicles, seasonal vegetation variation, or disaster-induced appearance changes. Handling such dynamic factors will be the focus of future work.

\end{document}